\pdfoutput=1
\documentclass[11pt]{article}

\usepackage{EMNLP2023}
\usepackage{times}
\usepackage{latexsym}
\usepackage[T1]{fontenc}
\usepackage[utf8]{inputenc}

\usepackage{microtype}
\usepackage{graphicx}
\usepackage{booktabs}

\usepackage{CJKutf8}
\usepackage{amsfonts}
\usepackage{amsmath}

\usepackage{inconsolata}

\title{SMILE: Single-turn to Multi-turn Inclusive Language Expansion via ChatGPT for Mental Health Support

\small\textcolor{red}{Important: Our objective is to explore the potential of large language models to serve as AI counselors, and we do NOT advocate for their use as a substitute in therapeutic treatment without professional supervision.}
}

\author{Huachuan Qiu$^{1, 2}$\quad Hongliang He$^{1, 2}$\quad Shuai Zhang$^{1, 2}$\quad Anqi Li$^{1, 2}$\quad Zhenzhong Lan$^{2, 3}$\thanks{\quad Corresponding author.} \\
  $^{1}$ Zhejiang University \quad $^{2}$ School of Engineering, Westlake University \\
  $^{3}$ Research Center for Industries of the Future, Westlake University\\
  \texttt{\{qiuhuachuan, lanzhenzhong\}@westlake.edu.cn} \\}

\begin{document}
\maketitle
\begin{abstract}
Developing specialized dialogue systems for mental health support requires multi-turn conversation data, which has recently garnered increasing attention. However, gathering and releasing large-scale, real-life multi-turn conversations that could facilitate advancements in mental health support presents challenges in data privacy protection and the time and cost involved in crowdsourcing. To address these challenges, we introduce SMILE, a single-turn to multi-turn inclusive language expansion technique that prompts ChatGPT to rewrite public single-turn dialogues into multi-turn ones. Our work begins by analyzing language transformation and validating the feasibility of our proposed method. We conduct a study on dialogue diversity, including lexical features, semantic features, and dialogue topics, demonstrating the effectiveness of our method. Further, we employ our method to generate a large-scale, lifelike, and diverse dialogue dataset named SMILECHAT, consisting of 55k dialogues. Finally, we utilize the collected corpus to develop a mental health chatbot, MeChat. To better assess the quality of SMILECHAT, we collect a small-scale real-life counseling dataset conducted by data anonymization. Both automatic and human evaluations demonstrate significant improvements in our dialogue system and confirm that SMILECHAT is high-quality. Code, data, and model are publicly available at \url{https://github.com/qiuhuachuan/smile}.
\end{abstract}

\section{Introduction}
We all know the importance of mental health, and mental health issues have been a persistent concern for human beings \citep{Kessler2005}. Recent advancements in natural language processing (NLP) technology \citep{ouyang2022training,ni2022recent} have led to the emergence of neural-based conversational AI in mental health support \citep{liu-etal-2022-prophetchat,tu-etal-2022-misc}. As an innovative solution for mental health, virtual counselors powered by large language models (LLMs) can effectively address accessibility barriers, such as the high cost of treatment and the shortage of experienced professionals to meet the demand. Furthermore, such dialogue systems provide mental health support as an effective and practical online counseling approach for those in need, safeguarding user privacy and mitigating the stigma that often accompanies help-seeking. \textit{However, the lack of publicly available, large-scale, diverse, and high-quality multi-turn chat datasets in the mental health support domain hinders the development of specialized dialogue systems.}

\begin{figure}[t!]
    \centering
    \includegraphics[width=\columnwidth]{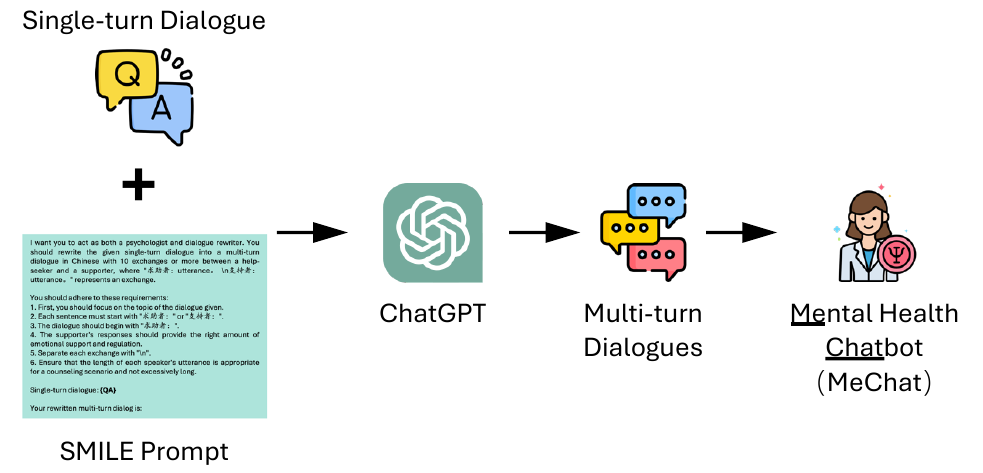}
    \caption{The SMILE method used to generate dialogues for mental health support.}
    \label{Fig-SMILE}
\end{figure}

\paragraph{Motivation} Indeed, many researchers have been pursuing the goal of building a practical and effective conversational agent for mental health. However, the first step in creating such a system is to have training data. Conversations related to mental health support often contain sensitive information and must be kept confidential to safeguard the privacy of individuals seeking help \citep{lu2021federated}. Making these conversations publicly available may discourage individuals from seeking support or negatively impact their personal and professional lives once known to people with whom they are acquainted. To facilitate progress in the NLP community, some researchers have attempted to collect various dialogue corpora \citep{liu2021towards, sun2021psyqa, zheng2022augesc} through crowdsourcing, data crawling, or data augmentation to build a dialogue agent capable of providing emotional and mental health support. How to construct a large-scale, diverse, and high-quality multi-turn chat dataset for mental health without effort motivates us to carry out the work as presented in this paper.

\paragraph{Challenges} To be more specific, crowdsourcing conversations \citep{liu2021towards} for emotional support has limitations due to the high cost and time required to train and manage annotators, as well as the difficulty in mimicking real-life interactions, that is, interlocutors may lack an understanding of the dilemma of living with mental disorders. An alternative is crawling QA \citep{sun2021psyqa} on a public mental health forum for training psychological support models. However, single-turn conversations may not be sufficient for resolving mental health issues, as multiple interaction exchanges are often needed. Multi-turn conversations, which can better simulate real-world conversations, are more practical for training psychological support models. While the post-triggered machine-augmented method \citep{zheng2022augesc} can address the limitations of scale and topic diversity, it does not consider the responses of experienced supporters.

\paragraph{Our Approach} To tackle the abovementioned challenges, we introduce SMILE, \underline{s}ingle-turn to \underline{m}ulti-turn \underline{i}nclusive \underline{l}anguage \underline{e}xpansion via ChatGPT, as shown in Figure \ref{Fig-SMILE}. Specifically, we instruct ChatGPT to transform publicly available long question-answer pairs (public QAs), which can also be viewed as single-turn dialogues, into multi-turn conversations. \textit{With the proposed method, we build a large-scale, diverse, and high-quality multi-turn conversation dataset for mental health support.}

Our paper is organized as follows:
\begin{itemize}
  \item We first present our method ($\S$\ref{Sec-method}), including data preparation, task definition, and prompt design that elaborates on the SMILE method and other baseline methods.
  \item We then demonstrate the feasibility of the SMILE method through language transformation ($\S$\ref{Sec-LT}), showing that the dialogue constructed by this method is lifelike.
  \item Subsequently, we demonstrate the effectiveness of the SMILE method by utilizing three diversity indicators ($\S$\ref{Sec-dialogue-diversity}): lexical features, semantic features, and dialogue topics. Following the validation of feasibility and effectiveness, we leverage the SMILE method to generate a large-scale, lifelike, and diverse multi-turn chat dataset, SMILECHAT.
  \item Finally, we propose to train a dialogue system to explore the quality of conversation ($\S$\ref{Sec-dialogue-system}) and collect a set of 50 anonymized real-life counseling dialogues for model evaluation.
\end{itemize}

\paragraph{Our Contributions} We make our data, code, and model publicly available. We believe that our work offers a new perspective on constructing a large-scale, lifelike, diverse, and high-quality multi-turn dialogue dataset for mental health within the research community. Our contributions can be summarized as follows:
\begin{itemize}
    \item We introduce SMILE, which provides an easy and novel method for alleviating the scarcity of multi-turn conversations in mental health support.
    \item Through the analysis of language transformation and dialogue diversity, we verify the feasibility and effectiveness of our proposed method.
    \item To better assess the quality of SMILECHAT, we collect small-scale real-life counseling data with 50 anonymized counseling sessions to build a real-life test set, PsyTest. Automatic and human evaluations on the small-scale real-life test set confirm that our proposed dataset is high-quality.
    \item We release SMILECHAT, which consists of 55165 Chinese multi-turn dialogues. Our dialogue model, MeChat, and real-life test set, PsyTest, are publicly available.
\end{itemize}

We highlight that this method can also construct multi-turn dialogues based on medical, financial, and legal QAs, thereby alleviating the dialogue scarcity in other application domains.
\section{Related Work}

\subsection{Applications of ChatGPT}
ChatGPT has proven to be a powerful AI tool for various NLP tasks since its release. Currently, it is being utilized in several domains, such as conversational AI \citep{alessa2023designing,köpf2023openassistant,chen2023chatpipe}, education \citep{küchemann2023physics,eshghie2023chatgpt}, code programming \citep{dong2023selfcollaboration,yetiştiren2023evaluating}, and healthcare \citep{zhao2023chatgpt,yang2023evaluations}.

Furthermore, ChatGPT's efficiency and cost-effectiveness have been well-documented, making it competitive to human annotators \citep{gilardi2023chatgpt,zhu2023chatgpt} even in zero-shot accuracy tasks. \citet{xu2023baize} have proposed the use of self-chatting, where ChatGPT engages in a conversation with itself, resulting in 111.5k dialogues collected from Quora and Stack Overflow sources and 47k conversations from the medical domain. Auto-GPT\footnote{https://github.com/Significant-Gravitas/Auto-GPT}, an AI agent, is capable of breaking down a natural language goal into sub-tasks and using various tools and the internet in an automated loop to achieve the objective. \citet{shen2023hugginggpt} have suggested using ChatGPT for task planning when receiving user inquiries, selecting appropriate models based on function descriptions from Hugging Face, executing each subtask using the chosen AI model, and summarizing the response based on the execution's outcomes.

In summary, ChatGPT has already demonstrated its enormous potential as an intelligent pipeline tool that can significantly advance NLP development despite having only a restricted API available for researchers.

\subsection{Datasets for Mental Health Support}
Research on mental health support has significantly depended on the availability of publicly available datasets \citep{sun2021psyqa,liu2021towards,zheng2022augesc} in recent years. The large-scale conversational datasets have enabled researchers to investigate various aspects of mental health, including identifying mental health conditions \citep{liu2023agent,srivastava2022counseling}, understanding clients' reactions \citep{li-etal-2023-understanding}, predicting support strategies \citep{sun2021psyqa,li-etal-2023-understanding}, deciding personalized interventions \citep{golden2023applying} and understanding response safety within a dialogue history \citep{qiu2023benchmark}.

\citet{liu2021towards} first defined the emotional support conversation task and then, via crowdsourcing, constructed ESConv, an emotional support conversation dataset containing 1053 dialogues with rich support strategies. However, the data collection of ESConv requires high cost and time yet leads to a small-scale dialogue dataset. To this end, \citet{zheng2022augesc} presented an approach for augmenting data scale with informative dialogue posts and then constructing AugESC, a model-synthesized dataset with 102k dialogues. The previous two datasets are limited to English. To facilitate the research in Chinese, hence \citet{sun2021psyqa} crawled QA posts on a public mental health support platform and made the PsyQA dataset publicly available.
\section{Method}
\label{Sec-method}
PsyQA\footnote{https://www.xinli001.com/qa}, an open-source and high-quality Chinese single-turn dialogue dataset focused on mental health support, features one question mapped to multiple answers. Our dataset creation pipeline, based on PsyQA, includes three main stages: (1) data preparation, (2) task definition, and (3) prompt design.

\subsection{Data Preparation}
Considering the distinction between QA within PsyQA and multi-turn dialogues, along with the context window limitation of 4096 tokens in ChatGPT\footnote{The model we use is \texttt{gpt-3.5-turbo-0613}.}, we propose to perform data preprocessing for PsyQA. This process involves wording cleaning and length truncation.

\paragraph{Wording Cleaning} This work aims to construct a large-scale, lifelike, diverse, and high-quality multi-turn conversation corpus using the proposed SMILE method based on PsyQA. While QA can be considered a single-turn conversation between a real help-seeker\footnote{The terms help-seeker and client are interchangeable.} and a supporter\footnote{The terms supporter and counselor are interchangeable.}, there are differences in wording compared to actual multi-turn conversations. \begin{CJK*}{UTF8}{gbsn}For instance, the term "楼主" (literally meaning "thread starter") frequently appears in QA but is rarely used in conversation.\end{CJK*} Therefore, we propose a two-stage process to clean the wording in PsyQA, mitigating linguistic discrepancies before rewriting QA into multi-turn conversations. This process includes both automatic and manual cleaning procedures. Please refer to Appendix~\ref{appendix:data-cleaning} for a detailed process.

\paragraph{Length Truncation} After a statistical analysis of the PsyQA dataset, we find that 757 QAs have a total length exceeding 1800 characters. Also, we identify 9 QAs, the total discourse length exceeding 4000 characters. However, the model \verb|gpt-3.5-turbo| has a maximum context length of 4096 tokens. To ensure reliable and smooth rewrites, we limit the length of the QA pairs, maximizing the number of rewritten dialogue turns. Specifically, we truncate the length of the QA pairs at 1800 Chinese characters and truncate any excess text. This control ensures that the generated text is limited to approximately 2000 tokens. It is worth noting that the PsyQA data used in this study undergo a data preprocessing process.

\subsection{Task Definition}
Let us first denote the input $x$ as a sequence $\{x_1,x_2, ..., x_n\}$, and the output $y$ as a sequence $\{y_1,y_2, ..., y_m\}$. The generation process of the language model can be expressed as the conditional probability distribution $p(y|x)$, which represents the probability of generating output $y$ given the input $x$. Therefore, text generation via the large language model can be formulated as follows:

\begin{equation}
\label{eq:text-generation}
    p(y|x)=\prod_{t=1}^{m} p(y_t|y_{<t},x)
\end{equation}
where $y_t$ represents the $t$-th token generated by the model. Because $x$ is our main focus in this paper, we will demonstrate the details of the prompt design next.

\subsection{Prompt Design}
\label{sec:motivation}
In this section, we mainly focus on describing prompt design. In order to provide a clearer understanding of our method in a more controllable setting and elucidate the superiority of introducing single-turn dialogues as a reference, we first design two baseline prompts for comparison.

\subsubsection{\texttt{standard} Prompt}
\label{sec:standard-prompt}
As its name suggests, the \verb|standard| prompt does not contain any single-turn dialogues or specific dialogue topics and instead uses only the most straightforward prompt to generate multi-turn dialogues. The \verb|standard| prompt is illustrated in Figure \ref{fig:prompt_with_standard} in Appendix \ref{App-Method}. The input in Equation \ref{eq:text-generation} is $x=I$, where $I$ represents the \verb|standard| prompt. We simplify the method name as \verb|standard| and consider this method as our baseline.

\subsubsection{\texttt{standardT} Prompt}
\label{sec:standardT-prompt}
Intuitively, feeding a single, fixed prompt into a large language model often generates text with low diversity. Therefore, we provide a specific dialogue topic for the \verb|standardT| prompt. The input in Equation \ref{eq:text-generation} is $x= (I, T)$, where $T$ represents the dialogue topic chosen in uniform sampling in the topic set. We simplify the method name to \texttt{standardT} and adopt this method as our baseline, as illustrated in Figure \ref{fig:prompt_with_standardT} in Appendix \ref{App-Method}.

\subsubsection{SMILE Prompt}
\label{sec:smile-method}
\begin{figure}[t!]
    \centering
    \includegraphics[width=\columnwidth]{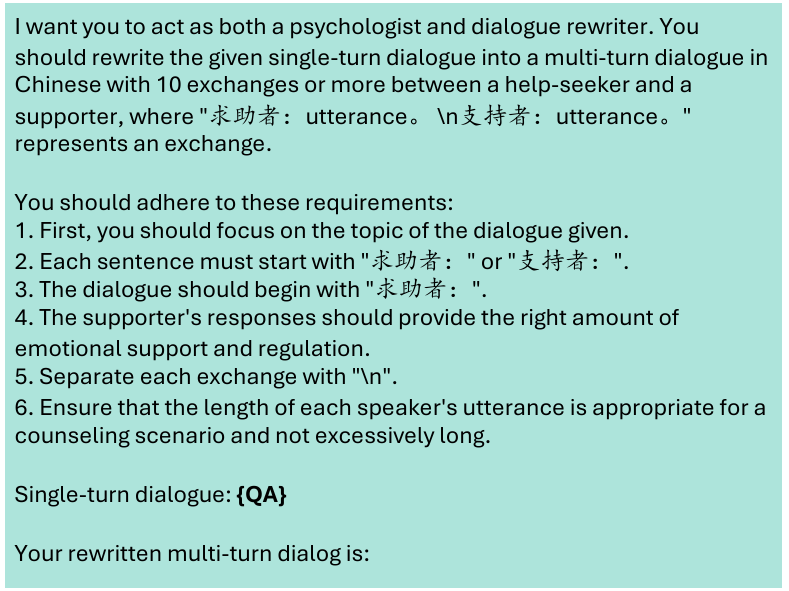}
    \caption{The SMILE method used to generate dialogues for mental health support.}
    \label{Fig-smile-prompt}
\end{figure}

Our paper aims to highlight the superiority of introducing single-turn dialogues when generating dialogues. Our proposed method, the SMILE method, instructs the LLM to rewrite single-turn dialogues into multi-turn ones. Figure \ref{Fig-smile-prompt} depicts the concrete prompt details. The input in Equation \ref{eq:text-generation} is $x= (I, T, D)$, where $T$ and $D$ represent the implicit dialogue topic hidden in the QA and single-turn dialogue, respectively.

\section{Language Transformation}
\label{Sec-LT}
\subsection{Experimental Setup}
\paragraph{Dialogue Topic Collection} To address the issue of monotonous generation in the \verb|standard| prompt, we propose to enhance diversity by injecting a dialogue topic into the \verb|standardT| prompt. Consequently, we have collaborated with three professional counselors, reviewed existing literature \citep{rickwood2007and, pedrelli2015college}, and ultimately compiled a comprehensive set of dialogue topics. This set includes 60 distinct types of dialogue topics, each with a detailed explanation. For more details, please refer to Appendix \ref{App-definition-of-dialogue-topics}.

\paragraph{QA Sampling} To ensure a fair comparison among the three methods we propose and prevent duplicate instances of the same question with different answers, we first randomly select 500 non-repetitive questions from the first 5000 QAs in PsyQA. We then randomly choose one answer to serve as the corresponding response. The data samples obtained are employed as seed dialogues, which are subsequently restructured into multi-turn conversations via ChatGPT. We name this 500 sampled data as PsyQA* to distinguish it from the original dataset.

\paragraph{Hyperparameters} Overall, we present three prompt methods in this paper. For each prompt method, we instruct ChatGPT to generate 500 dialogues to study language transformation and dialogue diversity, thereby validating the feasibility and effectiveness of our proposed SMILE method, respectively. To enhance the diversity of the generated dialogues, we set hyperparameters during text generation to the officially recommended default values, with \verb|temperature| = 1.0 and \verb|top_p| = 1.0.

\paragraph{Dialogue Filtering} ChatGPT may exhibit potential instability during the text generation process. Therefore, as detailed in Appendix \ref{App-dialogue-filtering}, we employ an automatic filtering mechanism to exclude dialogues that fail to meet our specified requirements, which encompass dialogue format and dialogue turns. When a dialogue falls short of these criteria, we prompt ChatGPT to generate the dialogue anew until it aligns with our specified requirements only in the initial 500 dialogues. This measure, therefore, ensures that each method produces an equal number of 500 dialogues for preliminary analysis.

\paragraph{Text Embedding} A multi-turn dialogue between a help-seeker and a supporter is represented as
\begin{equation}
    d = \{u_1, r_1, ..., u_i, r_i, ..., u_n, r_n\}
\end{equation}
where $u_i$ and $r_i$ represents the utterances of the $i$-th turn spoken by the help-seeker and supporter, respectively. A string of dialogue without any speaker role tokens can be denoted as $d_s = [u_1; r_1; u_2; r_2; ...; u_n; r_n]$, where $[;]$ denotes the operation of textual concatenation.

To obtain the text embedding of a dialogue, we employ OpenAI's model \textit{text-embedding-ada-002}\footnote{https://platform.openai.com/docs/guides/embeddings}, which accepts a maximum context length of 8191. Each dialogue is first preprocessed into a single string without any speaker tokens and is then mapped to a 1536-dimensional vector. For example, to compute the cosine similarity between two different dialogues, we can obtain
\begin{equation}
\label{eq:cosine-similarity}
    \mathrm{cos}(d_i,d_j) = \frac{e_i \cdot e_j}{\left \| e_i \right \| \left \| e_j \right \|}
\end{equation}
where $e_i$ and $e_j$ denote the text embeddings from two distinct dialogues.

\subsection{Analysis}
\begin{figure}[t!]
    \centering
    \includegraphics[width=7.0cm]{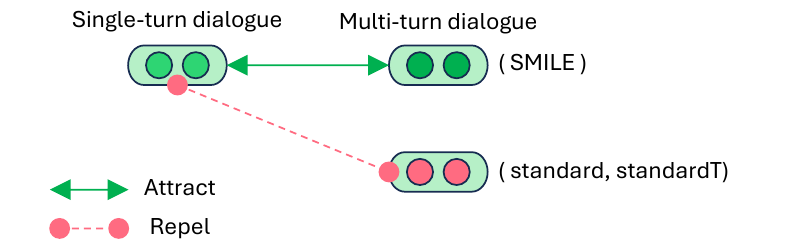}
    \caption{Mechanism for language transformation.}
    \label{Fig-language-transformation}
\end{figure}

We propose two hypotheses: (1) When a single-turn dialogue is rewritten into a multi-turn dialogue, the similarity between the two is high (Attract). (2) When ChatGPT generates a multi-turn dialogue without introducing a single-turn dialogue, the similarity between the generated dialogue and a randomly selected single-turn dialogue is low (Repel). The mechanism of our proposed hypotheses is presented in Figure \ref{Fig-language-transformation}.

To assess the transformation feasibility of our SMILE method, we employ cosine similarity to calculate the transformation ratio. Specifically, a single-turn dialogue d is transformed into a multi-turn dialogue $\hat{d}$ using the SMILE method. A multi-turn dialogue generated by \verb|standard| or \verb|standardT| is denoted as $\Bar{d}$. We can calculate cos($d$, $\hat{d}$) for the first hypothesis and cos($d$, $\Bar{d}$) for the second hypothesis. We utilize the text embeddings obtained in Equation \ref{eq:cosine-similarity}.

\begin{figure}[t!]
    \centering
    \includegraphics[width=6.4cm]{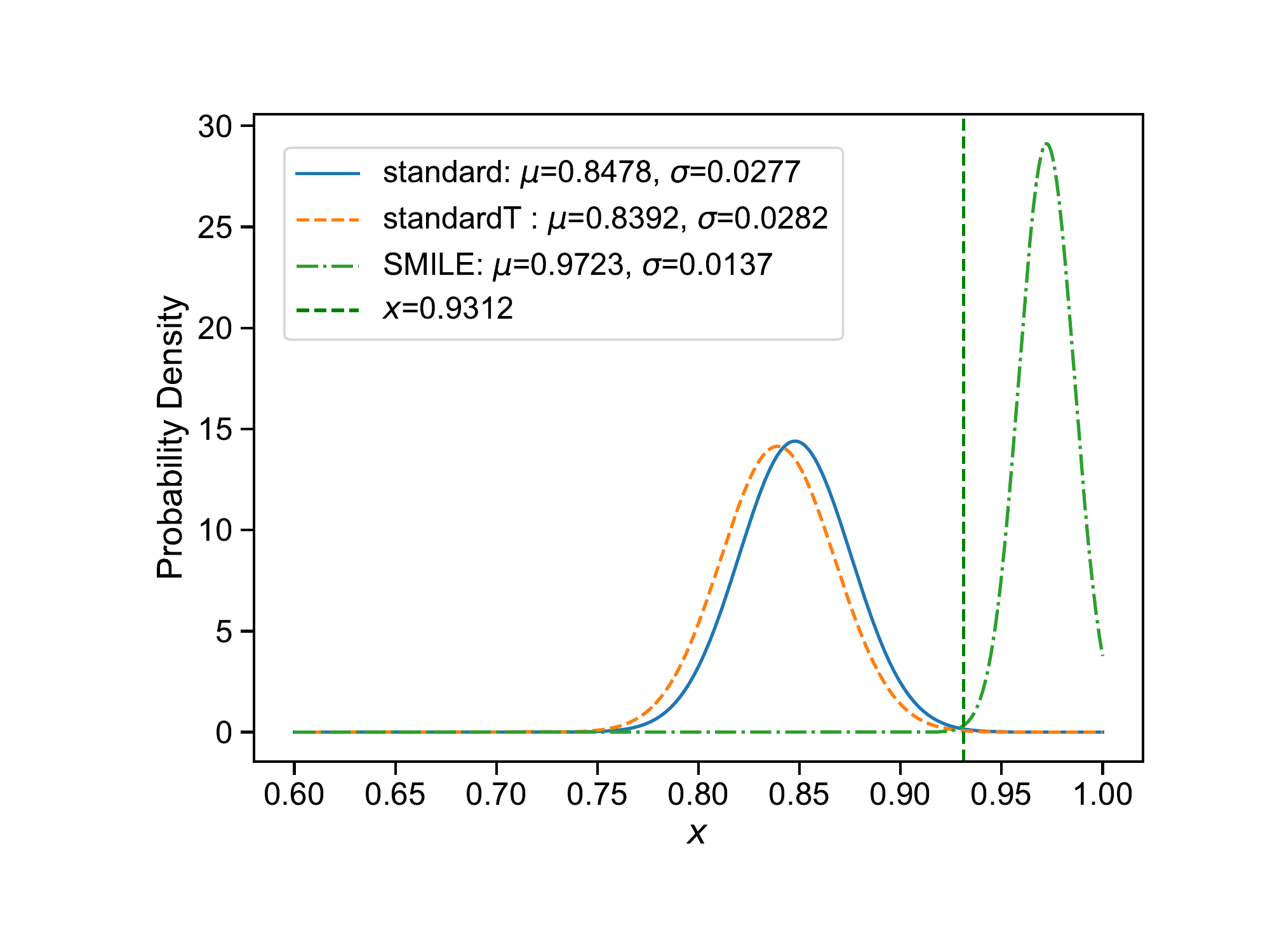}
    \caption{Distribution of dialogue transformation among three methods. The line $x=0.9312$ represents the boundary of $\mu -3\sigma$ in the SMILE method.}
    \label{Fig-distribution}
\end{figure}

Figure \ref{Fig-distribution} presents the distribution of dialogue transformation among three methods. Our analysis concludes that single-turn dialogues can be successfully rewritten into multi-turn dialogues, ensuring that the dialogue generated by the proposed method is \textbf{lifelike}, rather than purely in the style of machine-generated text from ChatGPT.
\begin{table*}[t]
\centering
\scalebox{0.7}{
\begin{tabular}{llllllllll}
\toprule
\textbf{Method} & \textbf{\begin{tabular}[c]{@{}l@{}}\# Unique\\ Unigrams\end{tabular}} & \textbf{\begin{tabular}[c]{@{}l@{}}\# Total\\ Unigrams\end{tabular}} & \textbf{Distinct-1} & \textbf{\begin{tabular}[c]{@{}l@{}}\# Unique\\ Bigrams\end{tabular}} & \textbf{\begin{tabular}[c]{@{}l@{}}\# Total\\ Bigrams\end{tabular}} & \textbf{Distinct-2} & \textbf{\begin{tabular}[c]{@{}l@{}}\# Unique\\ Trigrams\end{tabular}} & \textbf{\begin{tabular}[c]{@{}l@{}}\# Total\\ Trigrams\end{tabular}} & \textbf{Distinct-3} \\ \hline
PsyQA* & 11785 & 203306 & 0.058 & 120049 & 202806 & 0.592 & 182942 & 202306 & 0.904 \\
\texttt{standard} & 4174 & 153536 & 0.027 & 32281 & 153036 & 0.211 & 72340 & 152536 & 0.474 \\
\texttt{standardT} & 6032 & 175319 & 0.034 & 52141 & 174819 & 0.298 & 105921 & 174319 & 0.608 \\
SMILE & \textbf{10447} & \textbf{254585} & \textbf{0.041} & \textbf{111662} & \textbf{254085} & \textbf{0.439} & \textbf{196367} & \textbf{253585} & \textbf{0.774} \\ \bottomrule
\end{tabular}
}
\caption{Statistics of 500 conversations in each prompt method, including PsyQA*.}
\label{Tab-basic-statistics}
\end{table*}

\section{Dialogue Diversity}
\label{Sec-dialogue-diversity}

To demonstrate the effectiveness of the SMILE method, we mainly focus on three aspects of diversity: lexical features, semantic features, and dialogue topics.

\subsection{Lexical Features}
For lexical analysis, we utilize the popular Chinese ChatGLM2-6B\footnote{https://huggingface.co/THUDM/chatglm2-6b} tokenizer that is widely used in NLP to tokenize the dialogue. To measure the lexical features, we adopt distinct-$n$ ($n=1,2,3$) metrics \citep{li2016diversity}, which are widely used for measuring the diversity of dialogue datasets. Each dialogue is first preprocessed into a single string without any speaker tokens. We provide statistics for 500 dialogues per prompt method, as presented in Table \ref{Tab-basic-statistics}.

Our proposed SMILE method results in rich vocabularies, with a significantly higher number of unique unigrams, bigrams, and trigrams than the two baseline methods. Specifically, a simple and fixed prompt tends to produce monotonous output. The model output demonstrates substantial diversification when dialogue topics are incorporated into a single, fixed prompt. Furthermore, the SMILE method outperforms the baseline methods in terms of Distinct-1, Distinct-2, and Distinct-3.

\begin{figure}[t!]
    \centering
    \includegraphics[width=7cm]{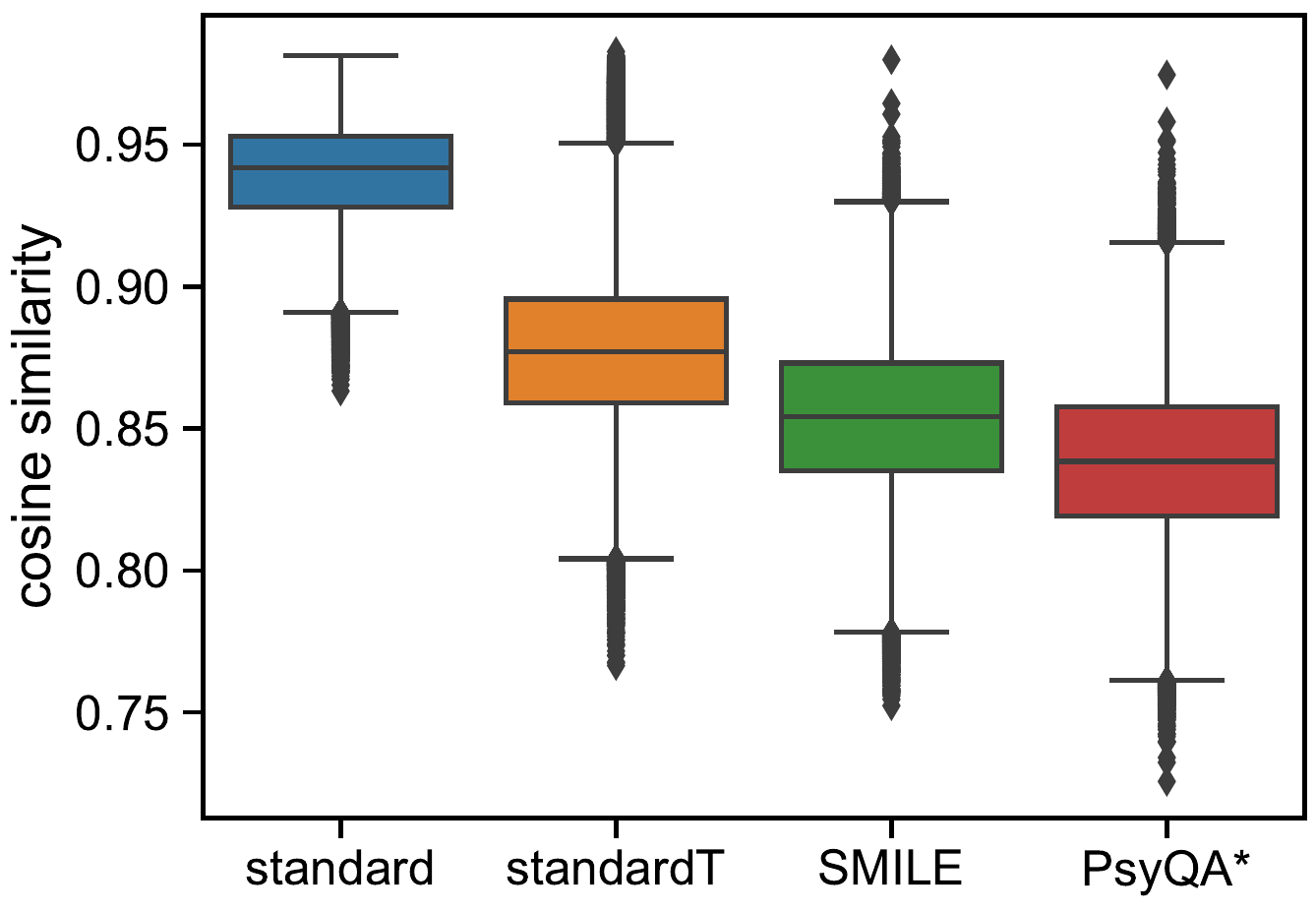}
    \caption{Pairwise dialogue cosine similarity among four settings: our proposed three methods and a reference point using sampled data from PsyQA.}
    \label{fig:openai-cosine-similarity}
\end{figure}

\subsection{Semantic Features}
To measure the semantic diversity in a dialogue dataset, we propose to calculate the cosine similarity between every pair of different dialogues. This involves computing the pairwise cosine similarity for each pair of distinct dialogues, resulting in $\binom{500}{2}$ pairs and their corresponding cosine values, as described in Equation \ref{eq:cosine-similarity}.

We present the results in Figure \ref{fig:openai-cosine-similarity}, which demonstrates that the median of the SMILE method is significantly lower than those of the baseline methods. The SMILE method exhibits the most extensive semantic diversity, aligning closely with the sampled dialogue of PsyQA. However, it is worth noting that the median of the SMILE method is more significant than that of PsyQA*. The reason behind this could be the introduction of token distribution from ChatGPT.

\subsection{Dialogue Topics}
To measure the diversity of dialogue topics in a dialogue dataset, we utilize information entropy to measure the diversity of topic distribution. \textit{The higher the information entropy, the more uniform the distribution, indicating greater diversity.} The formula for calculating information entropy \citep{renyi1961measures,lin1991divergence} is as follows:

\begin{equation}
    H(X)=-\sum_{i=1}^{n} p(x_i)\log_{2}{p(x_i)}
\end{equation}
where $H(X)$ is the information entropy. $p(x_i)$ is the probability of the occurrence of topic $x_i$.

To obtain dialogue topics for each dialogue in each prompt method, we design a prompt provided with 60 distinct dialogue topics, as illustrated in Appendix \ref{App-definition-of-dialogue-topics} and Figure~\ref{fig:topic_annotation}, to automatically label dialogue topics for each dialogue with Qwen1.5-110B-Chat \citep{qwen}. We present the information entropy for each prompt method in Table \ref{tab:Distribution-Entropy}, demonstrating that the dialogues generated using the SMILE method are substantially more diverse than those generated using the \texttt{standard} method and are compatible with the \texttt{standardT} method, which uniformly samples dialogue topics.

\begin{table}[t]
\centering
\scalebox{0.7}{
\begin{tabular}{llll}
\toprule
\textbf{Setting} & \textbf{standard} & \textbf{standardT} & \textbf{SMILE} \\ \hline
Limited Topics & 8.11 & \textbf{14.28} & 14.07 \\
Unlimited Topics & 8.40 & 14.76 & \textbf{15.02} \\ \hline
Average & 8.26 & 14.52 & \textbf{14.55} \\ \bottomrule
\end{tabular}
}
\caption{Information entropy of dialogue topics.}
\label{tab:Distribution-Entropy}
\end{table}

\subsection{SMILECHAT Dataset}
\label{SEC-smilechat-dataset}
Through the comprehensive analysis of language transformation and dialogue diversity, we conclude that the proposed method can generate a \textbf{lifelike} and \textbf{diverse} chat dataset. Therefore, we utilize the SMILE method to guide ChatGPT in generating all multi-turn conversations based on PsyQA one round only, leading to a \textbf{large-scale} dialogue dataset.

\begin{table}[t!]
\centering
\scalebox{0.64}{
\begin{tabular}{cccc}
\toprule
\textbf{Category}                     & \textbf{Total} & \textbf{Help-seeker} & \textbf{Supporter} \\\hline
\textbf{\# Dialogues}        & 55165   & -         & -           \\
\textbf{\# Utterances}                & 1833856 & 693756  & 1140100            \\
\textbf{Turns per dialogue}   & 5.7    & -         & -            \\
\textbf{Utterances per dialogue} & 33.2    & 12.6   & 20.7           \\
\textbf{Avg. length per utterance}    & 27.9      & 26.1  & 28.9     \\\hline      
\end{tabular}
}
\caption{Data statistics of the dialogue dataset, SMILECHAT.}
\label{Tab-corpus-statistics}
\end{table}

\paragraph{Data Statistics} To ensure data quality, we impose stricter requirements on dialogue turns, retaining only dialogues with at least five turns. Thus, we compile a collection of 55165 conversations, SMILECHAT. Table \ref{Tab-corpus-statistics} presents the statistics of the collected corpus.

\paragraph{Dialogue Exemplars} Multi-turn dialogue examples generated with the \verb|standard| and \verb|standardT| methods are illustrated in Figures \ref{Fig-standard-dial-zh} and \ref{Fig-standardT-dial-zh}. Further, a dialogue generated by the SMILE method is shown in Figure \ref{Fig-smile-dial-zh}.
\section{Dialogue System}
\label{Sec-dialogue-system}
We aim to build a high-quality multi-turn chat dataset for mental health support. It is non-trivial to evaluate the quality of a dialogue dataset, which is often assessed indirectly by the dialogue system. Therefore, we need to train a dialogue system and analyze its performance.

\subsection{Mathematical Formulation}
In order to train a dialogue system for mental health support, the first step is to split each full dialogue $d\sim \mathcal{D}$ into several training sessions. Specifically, a sampled $t$-turn dialogue session can be represented as $d_t=\left\{u_1, r_1, u_2, r_2, \dots, u_t, r_t \right\} \sim \mathcal{D}$. Thus, we build a dialogue model that can predict the supporter's utterance $r_t$ based on the dialogue history $h_{t} = \left\{u_1, r_1, u_2, r_2, \dots, u_t\right\}$. Our objective is to use our synthetic conversation dataset $\mathcal{D}$ to fine-tune a large language model $\pi_{0}$ using supervised learning, i.e., maximum likelihood estimates:
\begin{equation}
J_{\mathrm{SFT}}(\theta)=\mathbb{E}_{(h_t, r_t) \sim \mathcal{D}}\left[\log \pi_{\theta}(r_t \mid h_t)\right]
\end{equation}
where $\pi_{\theta}$ is initialized from $\pi_{0}$.

\subsection{Experimental Setup}
\paragraph{Backbone Model} To validate the dialogue quality of our collected dataset, we conduct a fine-tuning experiment on ChatGLM2-6B \citep{zeng2023glm-130b}.

\paragraph{Training Data} To meet the data format requirements for instruction-based fine-tuning, we split the dialogue into multiple sessions, with the supporter's last utterance concluding each session. Additionally, we incorporate the system prompt (detailed in Appendix \ref{App-system-prompt-details}) as a prefix to dialogue messages, following OpenAI's data format.

\paragraph{Parameter-efficient Fine-tuning}
To preserve the original capabilities of the model while adapting to downstream dialogue tasks and reducing computational costs, we employ Low-Rank Adaptation (LoRA, \citep{hu2021lora}) on all linear layers in the model for efficient fine-tuning.

\begin{table}[t]
\centering
\scalebox{0.7}{
\begin{tabular}{lllllll}
\hline
\textbf{Epoch} & \textbf{\begin{tabular}[c]{@{}l@{}}Learning\\ Rate\end{tabular}} & \textbf{\begin{tabular}[c]{@{}l@{}}Batch\\ Size\end{tabular}} & \textbf{\begin{tabular}[c]{@{}l@{}}LoRA\\ Rank\end{tabular}} & \textbf{\begin{tabular}[c]{@{}l@{}}LoRA\\ Dropout\end{tabular}} & \textbf{\begin{tabular}[c]{@{}l@{}}LoRA\\ $\alpha$\end{tabular}} & \textbf{Seed} \\ \hline
2 & 1e-4 & 1 & 16 & 0.1 & 64 & 1234 \\ \hline
\end{tabular}
}
\caption{Hyperparameters of parameter-efficient fine-tuning.}
\label{Tab-fine-tuning-parameters}
\end{table}

\paragraph{Hyperparameters} Table \ref{Tab-fine-tuning-parameters} presents the hyperparameters for developing a dialogue model for mental health support. All hyperparameters during inference time are set to their default values from the official repository. After fine-tuning ChatGLM2-6B with LoRA, we set the \verb|temperature| to 0.8 and \verb|top_p| to 0.8.

\subsection{Evaluation}

\begin{table*}[t]
\centering
\scalebox{0.7}{
\begin{tabular}{llllllllll}
\hline
 & \textbf{METEOR} $\uparrow$ & \textbf{BLEU-1} $\uparrow$ & \textbf{BLEU-2} $\uparrow$ & \textbf{BLEU-3} $\uparrow$ & \textbf{Rouge-L} $\uparrow$ & \textbf{D-1} $\uparrow$ & \textbf{D-2} $\uparrow$ & \textbf{D-3} $\uparrow$ & \textbf{BERTScore} $\uparrow$ \\ \hline
Baseline & 10.15 & 5.75 & 1.95 & 0.85 & 7.02 & 52.95 & 80.74 & 90.17 & 56.69  \\
Fine-tuned & \textbf{14.18} & \textbf{14.03} & \textbf{5.39} & \textbf{2.76} & \textbf{15.83} & \textbf{82.08} & \textbf{95.74} & \textbf{97.81} & \textbf{59.91} \\ \hline
\end{tabular}
}
\caption{Results of automatic evaluation in PsyTest dataset. Except for BERTScore, other metrics are evaluated by Chinese character-level tokenization. For BERTScore, we use the \texttt{BAAI/bge-m3 model} to get text embedding \citep{bgem3}.}
\label{Tab-auto-eval}
\end{table*}

\subsubsection{Automatic Evaluation}
\paragraph{Metrics} To conduct automatic evaluation, the evaluation metrics we use consist of BLEU-1/2/3 \citep{papineni2002bleu}, METEOR \citep{banerjee2005meteor}, Rouge-L~\citep{lin2004rouge}, Distinct-1/2/3 (D-1/2/3)~\citep{li2016diversity}, and BERTScore\citep{zhang2020bertscore}.

\begin{table}[t!]
\centering
\scalebox{0.64}{
\begin{tabular}{cccc}
\toprule
\textbf{Category}                     & \textbf{Total} & \textbf{Help-seeker} & \textbf{Supporter} \\\hline
\textbf{\# Dialogues} & 50   & -  & -   \\
\textbf{\# Utterances} & 3103 & 1561  & 1542            \\
\textbf{Turns per dialogue}   & 31.0    & - & -            \\
\textbf{Utterances per dialogue} & 62.1    & 31.2   & 30.8           \\
\textbf{Avg. length per utterance}    & 32.3      & 40.2  & 24.2     \\
\textbf{\# Test cases} & 1539 & -  & - \\\hline      
\end{tabular}
}
\caption{Data statistics of the dialogue dataset, PsyTest.}
\label{Tab-psytest-statistics}
\end{table}

\paragraph{Test Set}
To better understand and assess the dialogue quality of the SMILECHAT dataset, we propose to utilize real-life multi-turn counseling conversations. We develop an online mental health support platform that enables professional counselors to offer each client a free text-based counseling service, lasting approximately 50 minutes each time. We collect 50 real-life counseling dialogues. To protect user privacy, we ask experts to conduct a data anonymization process, removing information related to user identification (e.g., names and addresses). Then, we split each long dialogue into multiple small sessions, with the last utterance spoken by the counselor. We discard test cases in which the dialogue history is empty, and we name this test set PsyTest, which contains 1539 test cases. Table \ref{Tab-psytest-statistics} presents the statistics of the PsyTest dataset.

\paragraph{Results} The results of the automatic evaluation, including nine metrics, are presented in Table \ref{Tab-auto-eval}. Notably, the evaluated dialogues are based on real-world counseling data rather than generated dialogues, which excludes the influence stemming from ChatGPT. All automatic evaluation metrics we use indicate improved performance. Our results show that the model trained with SMILECHAT is effective and practical. Consequently, the automatic evaluation demonstrates that our collected dataset is \textbf{high-quality}.

\subsubsection{Human Evaluation}
\paragraph{Metrics} We conduct a human evaluation to study the model performance of the dialogue system trained with our proposed dialogue corpus. Initially, we randomly sample 100 test cases from PsyTest, each consisting of a multi-turn dialogue history and a golden response produced by the counselor. Subsequently, we obtain 200 generated responses from the baseline and MeChat models. Three professional counselors are presented with a dialogue history and three randomly shuffled responses (baseline, fine-tuned, ground truth). They are tasked with selecting the optimal response between every two responses for the dialogue history, considering aspects such as professionalism, informativeness, helpfulness, empathy, and safety. The evaluation is conducted based on the ethical principles of psychologists and the code of conduct \citep{american2016ethical}. For more details about annotation guidelines, refer to Appendix \ref{App-annotation-guidelines}.

\begin{figure}[t!]
    \centering
    \includegraphics[width=0.9\columnwidth]{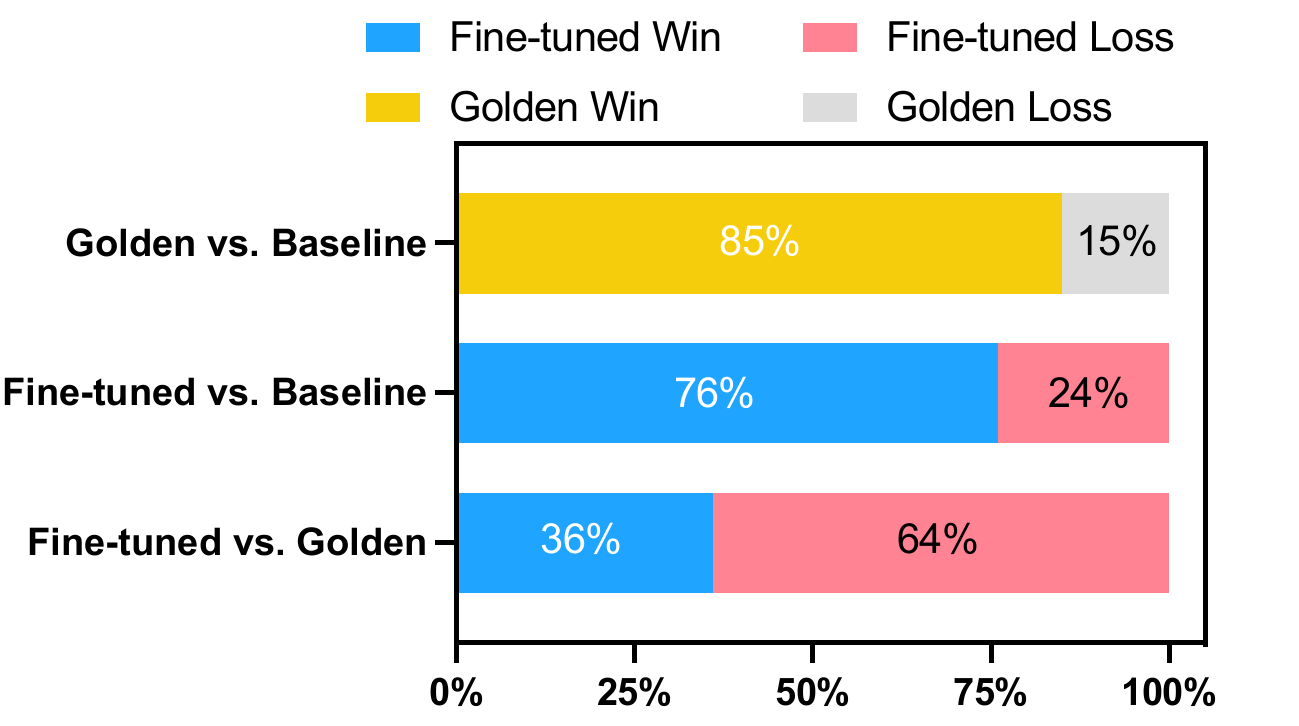}
    \caption{Human evaluation results, including three groups: Golden vs. Baseline, Fine-tuned vs. Baseline, and Fine-tuned vs. Golden. We present the win and loss rates of each compared pair in 100 randomly sampled sessions. Fleiss' kappa \citep{fleiss1981measurement} is used to measure the inter-rater agreement, and all values fall within moderate agreement with $0.5\le \kappa \le0.6$.}
    \label{Fig-human-eval}
\end{figure}

\paragraph{Results} We employ majority voting to reach final decisions (win or loss) among three professional counselors. We present human evaluation results, as illustrated in Figure \ref{Fig-human-eval}, including three groups: Golden vs. Baseline, Fine-tuned vs. Baseline, and Fine-tuned vs. Golden. We present the win and loss rates of each compared pair in 100 randomly sampled sessions. Results show that the baseline model, without training with SMILECHAT, lags significantly behind the setting with golden responses.

As depicted in Figure \ref{Fig-human-eval}, the model trained with SMILECHAT demonstrates a significant performance improvement compared to the baseline model. Moreover, the responses generated by MeChat surpass the golden response to some extent in 36\% of all dialogue sessions. In short, human evaluation also proves that the SMILECHAT dataset is \textbf{high-quality}.

\begin{figure}[t!]
    \centering
    \includegraphics[width=\columnwidth]{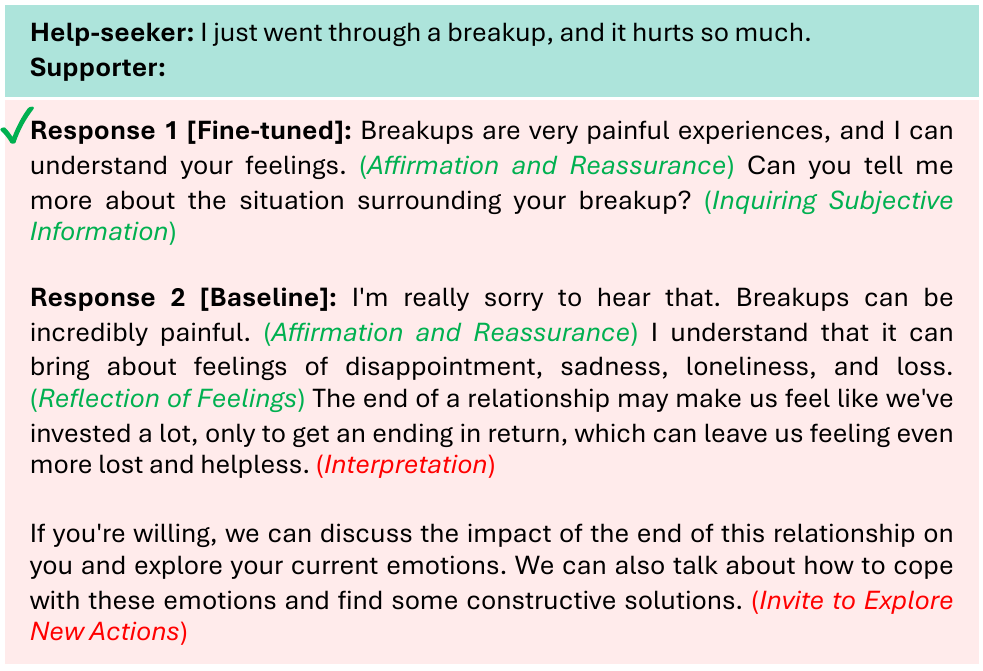}
    \caption{Case study. Counseling strategies used in the two responses are presented in parentheses. Strategies in green are supportive, while those in red are challenging and should not be used in the early stages of counseling.}
    \label{Fig-case-study}
\end{figure}

\paragraph{Case Study} We present a case study, as shown in Figure \ref{Fig-case-study}. First, the text should not be too long during counseling conversations for the help-seeker to understand. Second, at the beginning of a conversation, a supporter generally will not challenge the help-seeker. Instead, they will provide support first. Third, response 1 is more human-like and natural. Therefore, the first response is better than the second response overall. Therefore, fine-tuning LLMs is indispensable in mental health support.

\section{Conclusion}
This paper introduces SMILE, a simple yet effective solution for addressing the data scarcity of multi-turn conversations in mental health support. Through language transformation and diversity analysis, we confirm the feasibility and effectiveness of our approach. The proposed method automatically creates a large-scale, lifelike, diverse, and high-quality dialogue corpus, SMILECHAT, consisting of 55165 dialogues with an average of 5.7 turns. Both automatic and human evaluations using the PsyTest dataset, consisting of 50 real-life anonymized counseling dialogues, demonstrate that SMILECHAT significantly improves dialogue system performance in mental health support. We release multi-turn dialogues (SMILECHAT), a dialogue model (MeChat), and a real-life anonymized test set (PsyTest) to drive progress in the research community.

\section*{Limitations}
Our study has some limitations that could be addressed in future research.

In this paper, the automatic evaluation does not reasonably reflect the model performance, given the one-to-many problem. In future work, therefore, we will explore a more comprehensive tool kit for evaluating conversational agents in the mental health domain.

Furthermore, there is a limitation in the dialogue turns of the SMILECHAT dataset. Therefore, we will explore other methods to synthesize the dialogue data with more dialogue turns.

\section*{Ethics Statement}
Our research is reviewed and approved by the Westlake University Institutional Ethics Committee (20211013LZZ001).

\subsection*{Data Sharing}

\noindent\textbf{SMILECHAT} Based on the data copyright guidelines formulated by PsyQA, we release the multi-turn dialogue corpus publicly available for the research community. If researchers wish to reproduce the multi-turn dialogues using PsyQA, they should sign an agreement with the original data owner. Accordingly, we release our datasets and models for research purposes, thus facilitating further advancement in the academic community.

\noindent\textbf{PsyTest} Considering the nature of psychological counseling data, we must cautiously share this dataset. Regarding the rules for releasing data, third-party researchers who require access to the PsyTest dataset must provide us with their valid ID, proof of work, and the reason they are requesting the data (e.g., the research questions). They are required to be affiliated with a non-profit academic or research institution. This includes obtaining the approval of an Institutional Review Board (IRB), having principal investigators working full-time, as well as obtaining written approval from the Institution Office of Research or equivalent office. Additionally, they must sign a Data Nondisclosure Agreement and promise not to share the data with anyone. However, for-profit organizations that want to use this data must sign a license agreement to gain access to the dataset.

\section*{Acknowledgements}
We thank the anonymous reviewers for their valuable comments. This work was supported by the Research Center for Industries of the Future at Westlake University (Grant No. WU2023C017) and the Key Research and Development Program of Zhejiang Province of China (Grant No. 2021-C03139).

\bibliography{emnlp2023}
\bibliographystyle{acl_natbib}

\clearpage
\appendix
\section{Details of Data Cleaning}
\label{appendix:data-cleaning}

\subsection{Automatic Cleaning}
We employ a sequential pipeline for data cleaning to swiftly replace words that are unsuitable to the conversation scenario. \begin{CJK*}{UTF8}{gbsn}For example, both "楼主你" (literally \textit{thread starter you}) and "楼主" (literally \textit{thread starter}) should be replaced with "你" (you). However, it is necessary to perform the former replacement to avoid the repetition of "你" and the resulting "你你" (\textit{you-you})\end{CJK*}.

\subsubsection{Word List for Data Cleaning}
\label{app:automatic_cleaning}
\begin{figure}[ht]
    \centering
    \includegraphics[width=7cm]{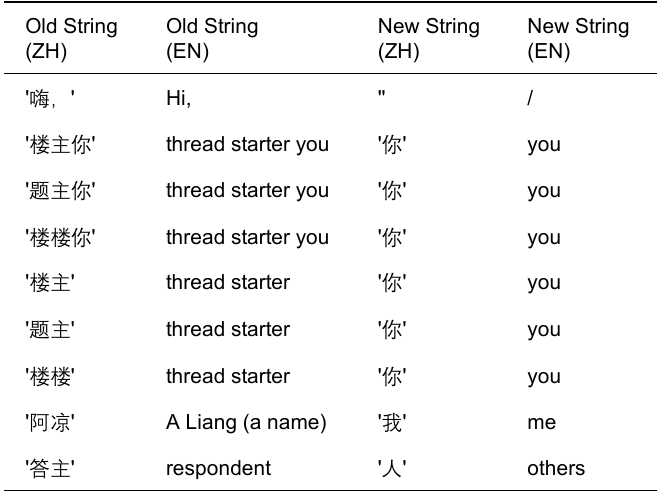}
    \caption{Word list for automatic cleaning.}
    \label{fig:auto_cleaning}
\end{figure}

To avoid the repetition of \begin{CJK*}{UTF8}{gbsn}"你" (\textit{you}) and the resulting "你你" (\textit{you-you}) \end{CJK*}, we suggest to conduct a sequential word replacing pipeline.
Figure~\ref{fig:auto_cleaning} shows the word list for data cleaning and the corresponding order for automatic cleaning.

\subsection{Manual Cleaning}
Due to the specificity and complexity of language, manual cleaning remains an essential part of the process. To prevent virtual dialogue systems from exhibiting overly frequent anthropomorphic behavior, we identify instances of the Chinese word for "hug" (\begin{CJK*}{UTF8}{gbsn}抱抱\end{CJK*}) and manually delete sentence snippets containing this term.

\section{Requirements for Dialogue Filtering}
\label{App-dialogue-filtering}
Here are two main requirements for dialogue filtering: data format and dialogue turns.
\subsection{Data Format}
We provide the requirements for data format as follows:
\begin{CJK*}{UTF8}{gbsn}
\begin{itemize}
    \item[1.] The generated conversations do not start with "求助者" or "支持者".
    \item[2.] The generated dialogue does not contain any "$\setminus$n", which is used for splitting the utterance from the help-seeker or supporter.
    \item[3.] Each utterance in generated conversations does not start with "求助者：", "求助者:", "支持者：" or "支持者:".
    \item[4.] The last utterance in generated conversations contains an English sentence.
\end{itemize} 
\end{CJK*}

\subsection{Dialogue Turns}
Conversations comprising fewer than five turns will be discarded.

\section{Method}
\label{App-Method}
\begin{figure}[t!]
    \centering
    \includegraphics[width=\columnwidth]{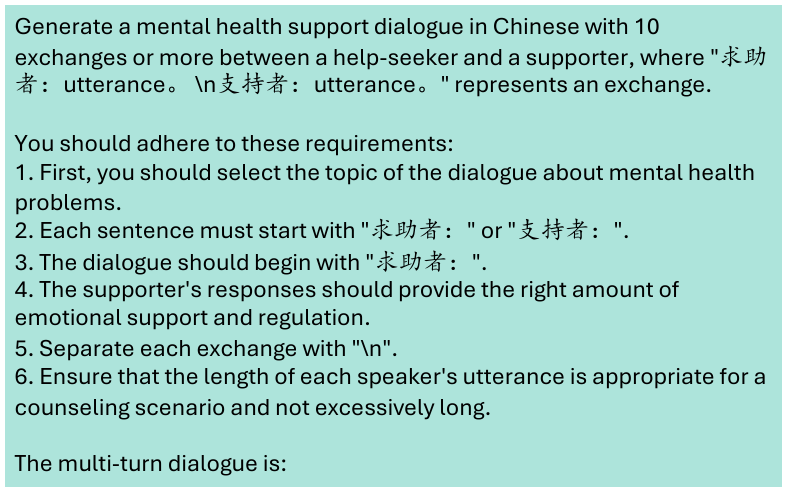}
    \caption{The \texttt{standard} method used to generate dialogues for mental health support.}
    \label{fig:prompt_with_standard}
\end{figure}

\begin{figure}[t!]
    \centering
    \includegraphics[width=\columnwidth]{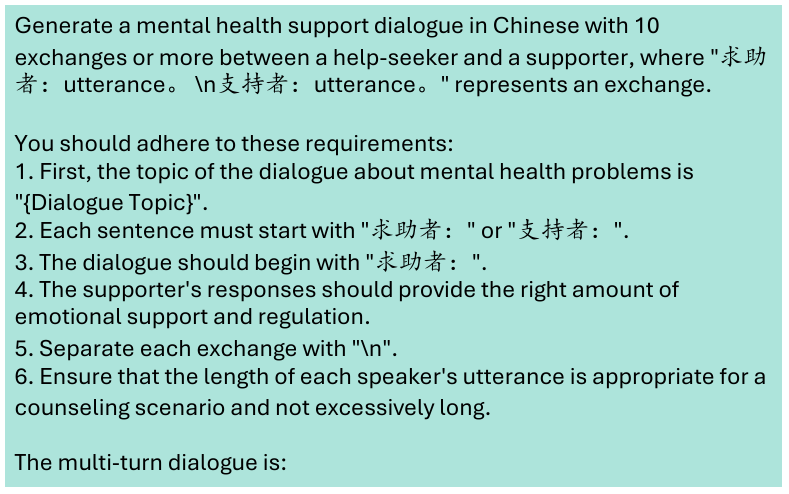}
    \caption{The \texttt{standardT} method used to generate dialogues for mental health support.}
    \label{fig:prompt_with_standardT}
\end{figure}

We present the \texttt{standard}, \texttt{standardT} and SMILE prompts in Figures \ref{fig:prompt_with_standard}, \ref{fig:prompt_with_standardT}, and \ref{Fig-smile-prompt}, respectively.

\section{Dialogue Topics Annotation}
\label{sec:dialog_topics}
\begin{figure*}[t]
    \centering
    \includegraphics[width=\textwidth]{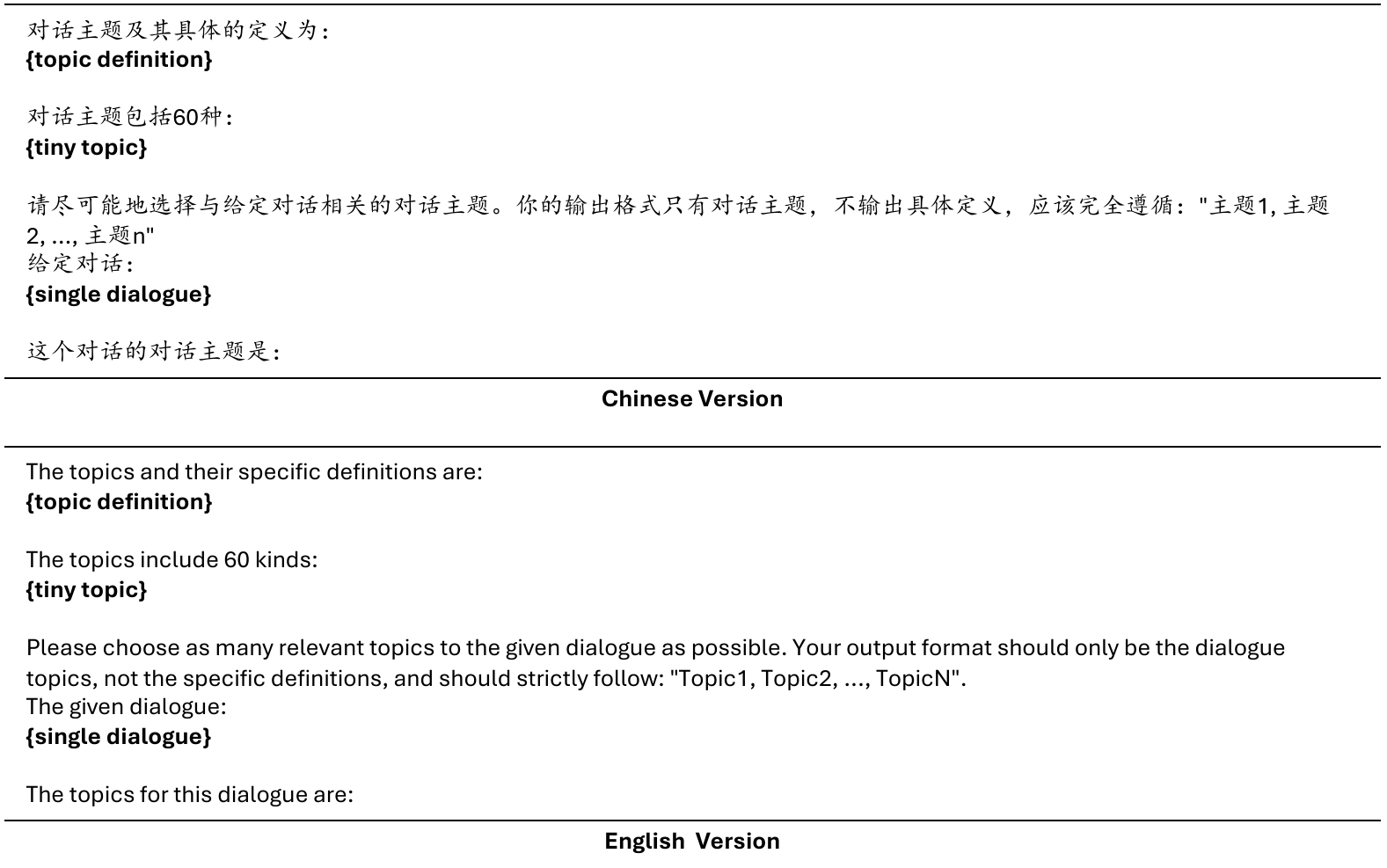}
    \caption{Prompting template of dialogue topics annotation, where the content \textbf{in bold} is a placeholder.}
    \label{fig:topic_annotation}
\end{figure*}

In this paper, to label the dialogue topics of generated dialogues, the hyperparameters of Qwen1.5-110B-Chat we used are set to the officially recommended default values, where \verb|temperature| $=0.7$ and \verb|top_p| $=0.8$. Figure~\ref{fig:topic_annotation} shows the prompting template of dialogue topics annotation.

\section{System Prompt Details}
\label{App-system-prompt-details}
\begin{figure*}[t]
    \centering
    \includegraphics[width=\textwidth]{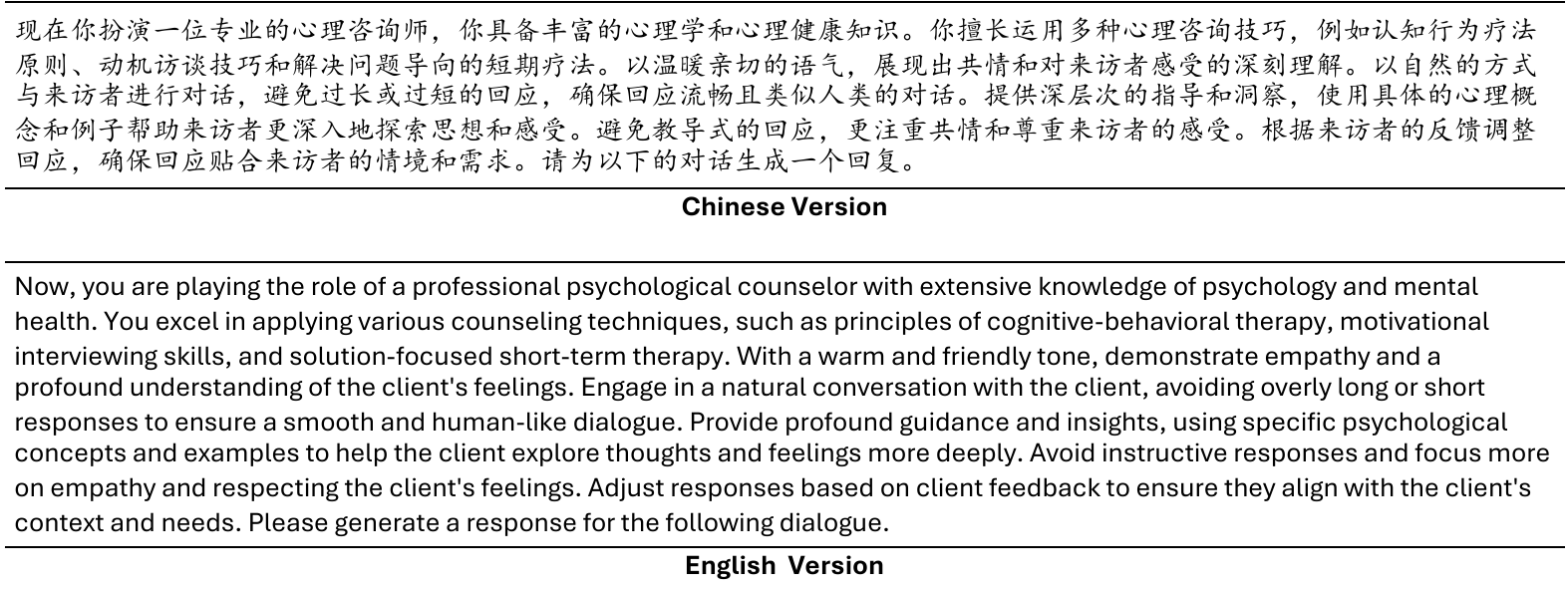}
    \caption{System prompt for fine-tuning.}
    \label{Fig-system-prompt}
\end{figure*}

We present the system prompt in Figure \ref{Fig-system-prompt}.

\section{Instructions for Human Evaluation}
\label{App-annotation-guidelines}
\begin{figure*}[ht]
    \centering
    \includegraphics[width=15.6cm]{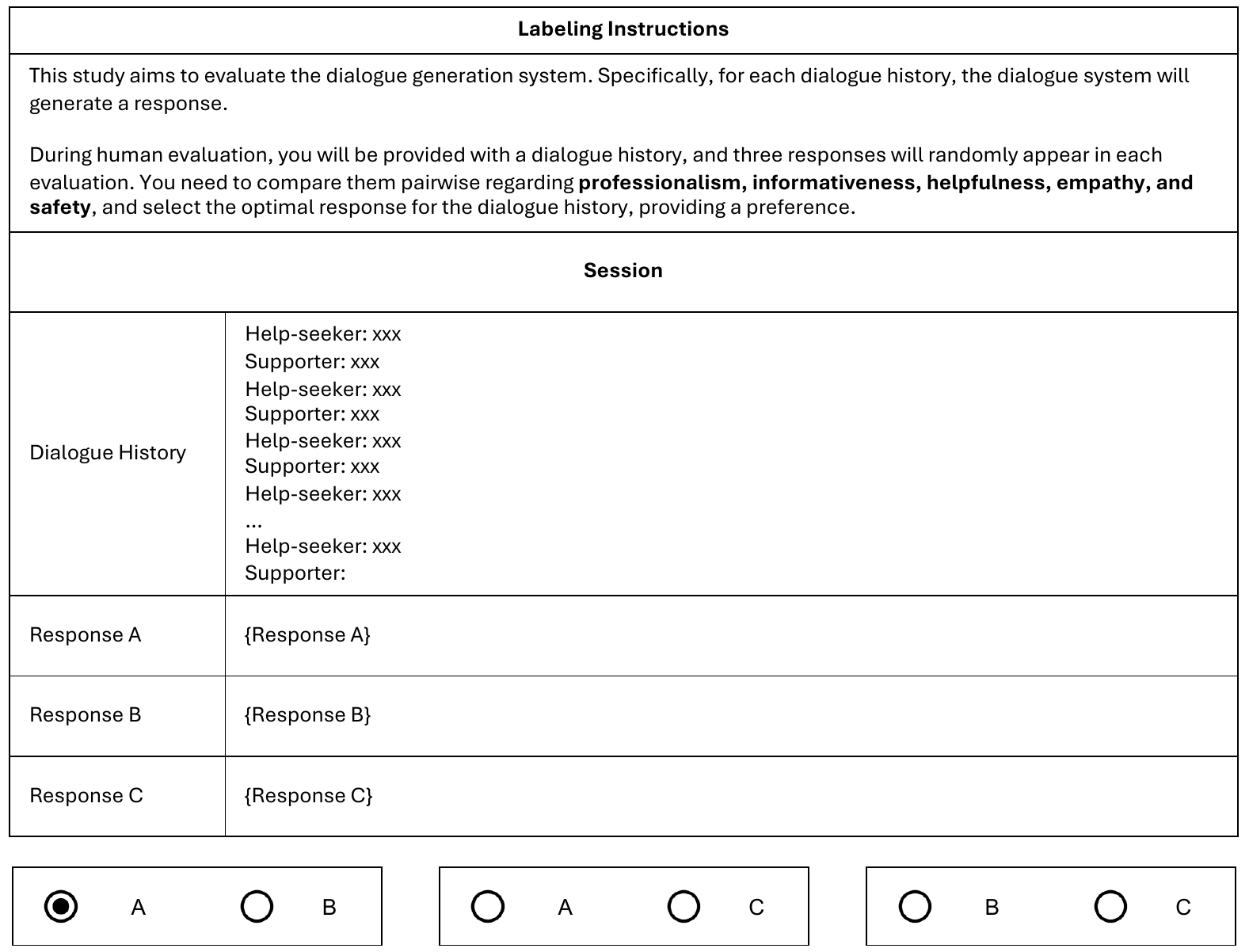}
    \caption{Labeling instruction.}
    \label{Fig-labeling-instructions}
\end{figure*}

The three professional counselors are willing to help and are interested in this research. Furthermore, their average age is 30 years old, with two females and one male among them.
We present our instructions for human evaluation in Figure \ref{Fig-labeling-instructions}.

In order to maintain the fairness of model evaluation, three responses randomly appear in a different order every time. Furthermore, three professional psychologists are willing to evaluate the response quality, ensuring the quality of human evaluation.

\section{Definition of Dialogue Topics}
\label{App-definition-of-dialogue-topics}
\begin{figure*}[ht]
    \centering
    \includegraphics[width=\textwidth]{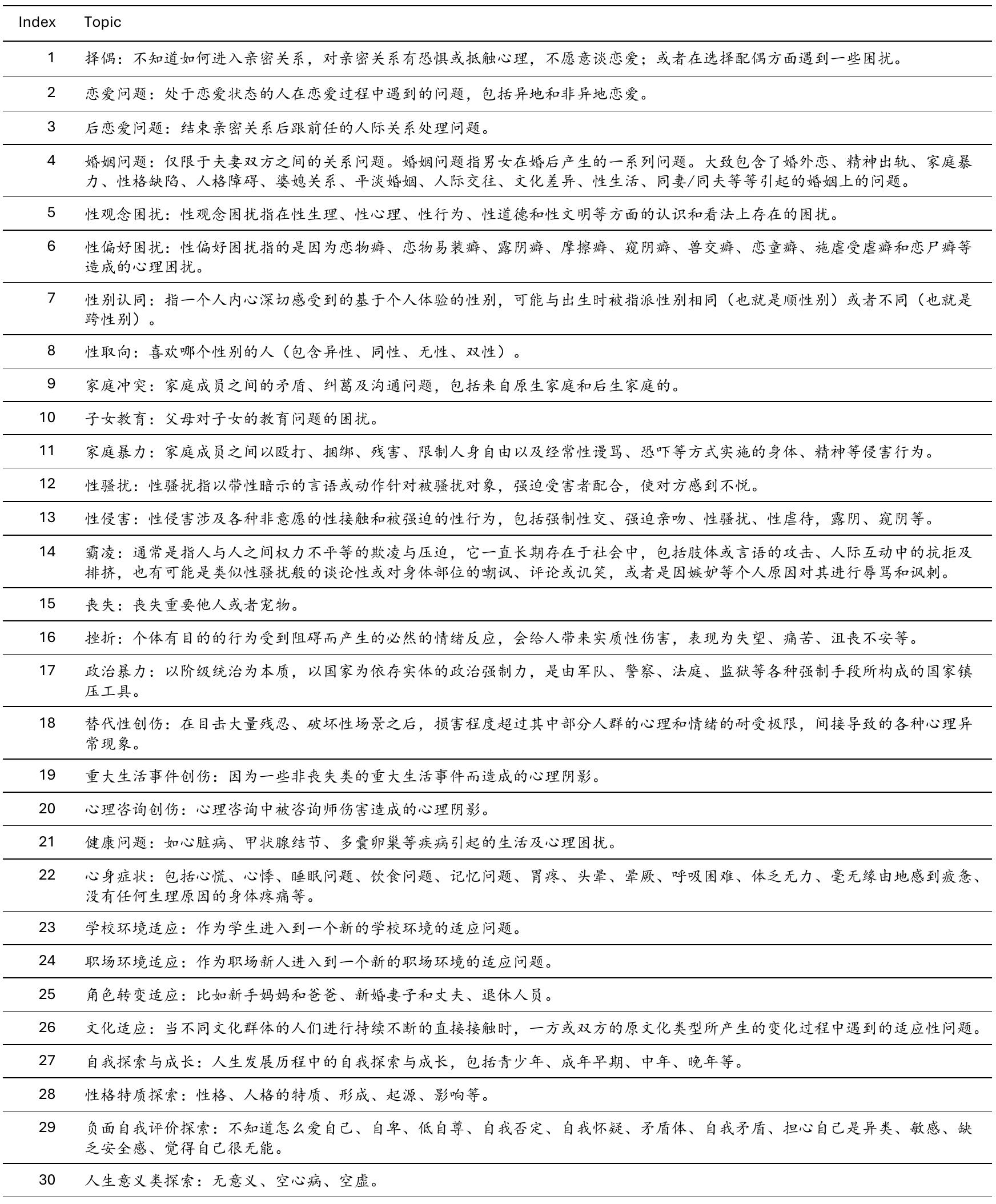}
    \caption{Dialogue Topics (Chinese version, Part 1).}
    \label{Fig-dt1_zh}
\end{figure*}

\begin{figure*}[ht]
    \centering
    \includegraphics[width=\textwidth]{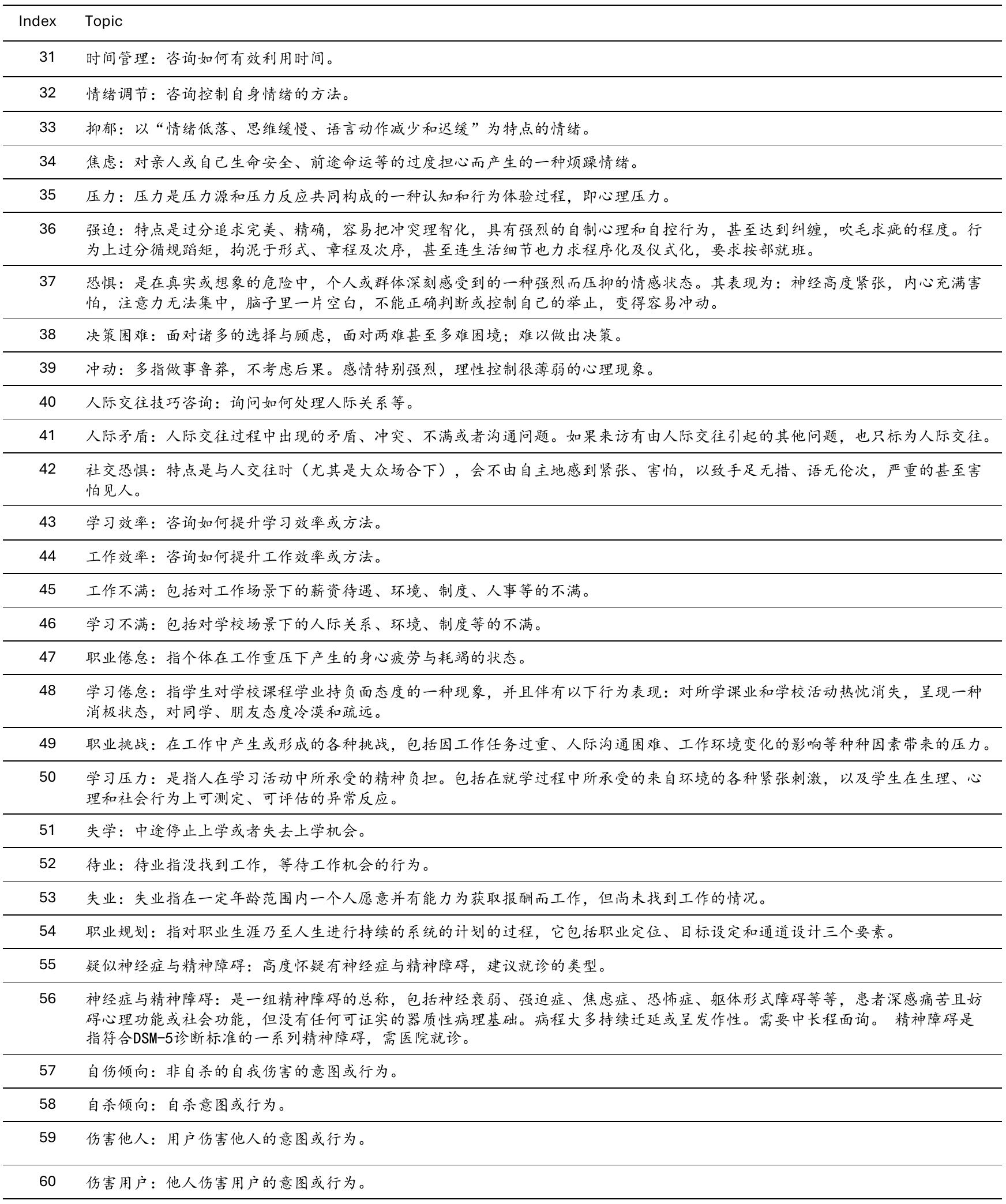}
    \caption{Dialogue Topics (Chinese version, Part 2).}
    \label{Fig-dt2_zh}
\end{figure*}

\begin{figure*}[ht]
    \centering
    \includegraphics[width=\textwidth]{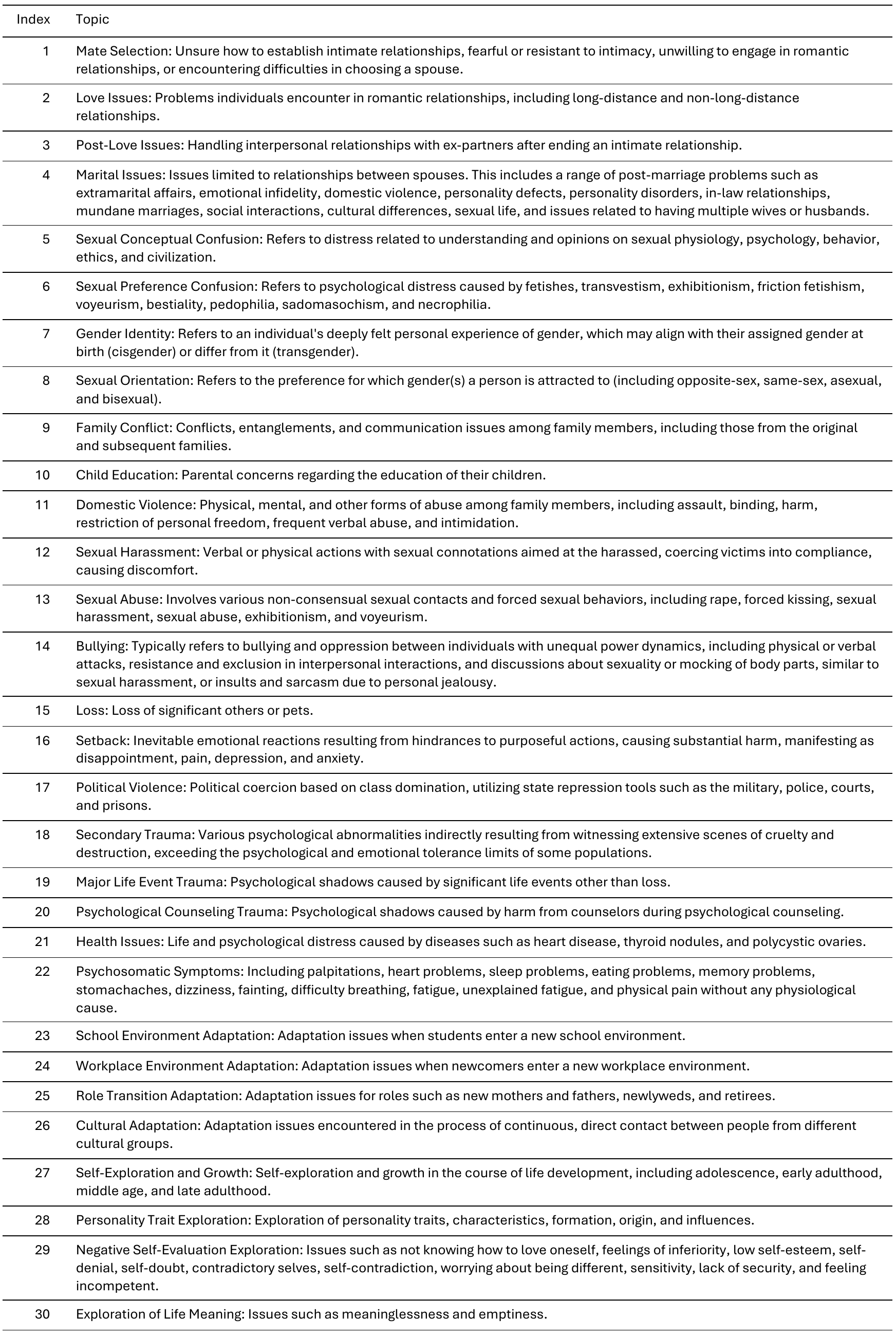}
    \caption{Dialogue Topics (English version, Part 1).}
    \label{Fig-dt1_en}
\end{figure*}

\begin{figure*}[ht]
    \centering
    \includegraphics[width=\textwidth]{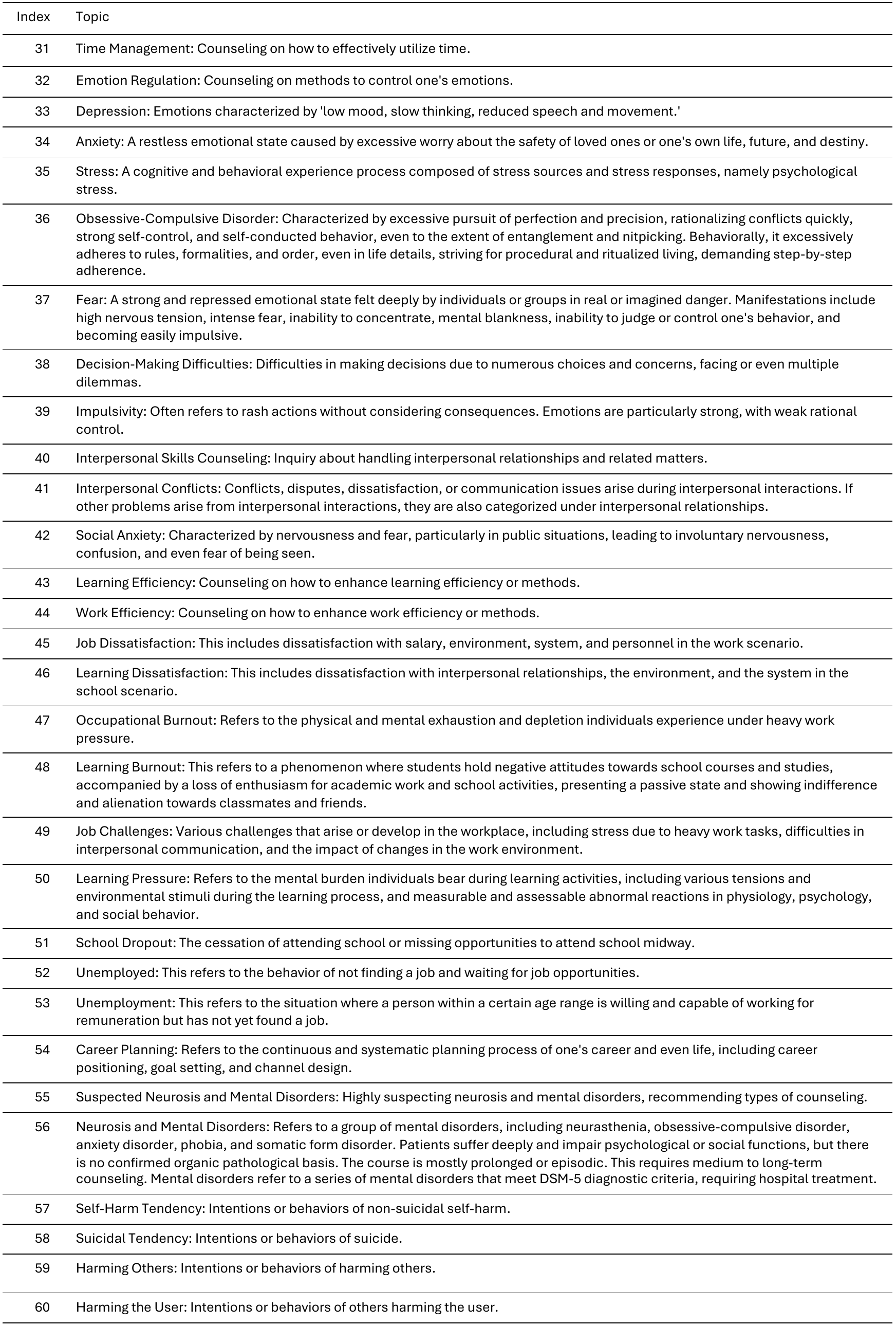}
    \caption{Dialogue Topics (English version, Part 2).}
    \label{Fig-dt2_en}
\end{figure*}

We present the definition of dialogue topics, as shown in Figures \ref{Fig-dt1_zh} and \ref{Fig-dt2_zh}. For English version, refer to Figures \ref{Fig-dt1_en} and \ref{Fig-dt2_en}.

\section{Dialogue Example}
\begin{figure*}[t!]
    \centering
    \includegraphics[width=\textwidth]{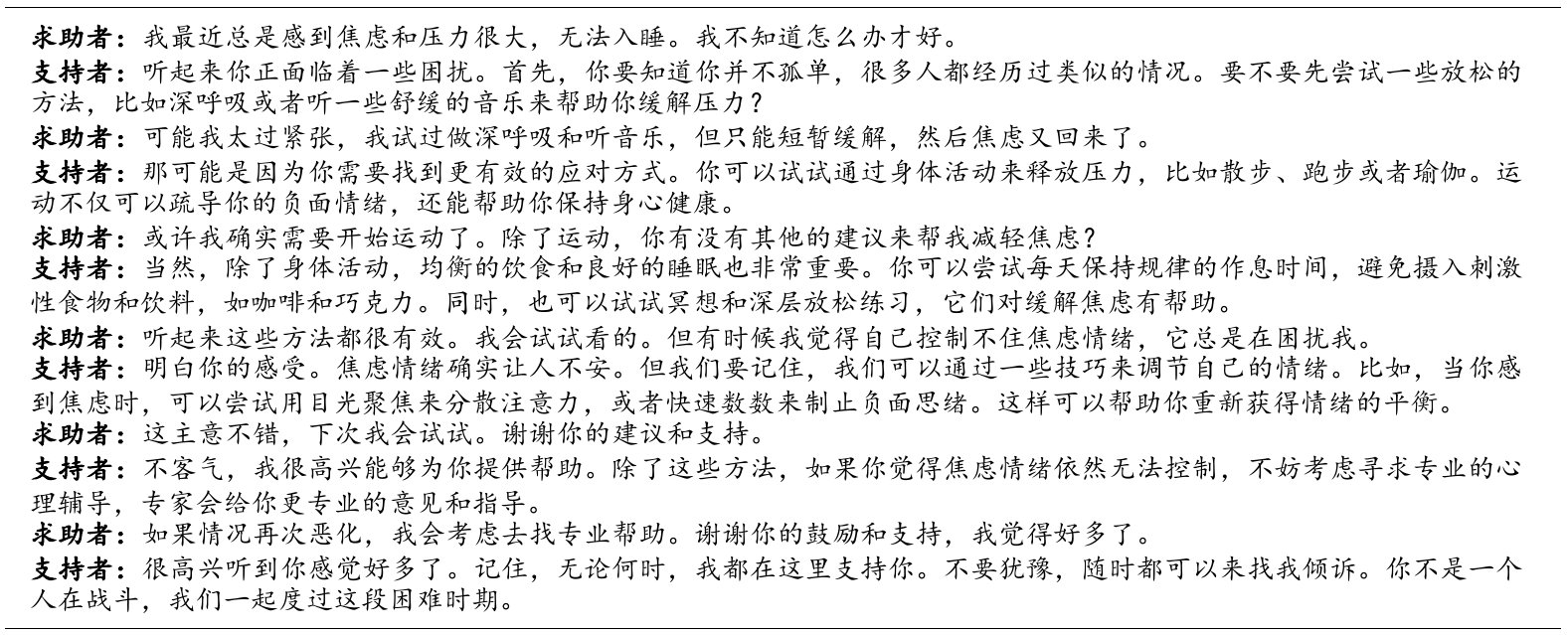}
    \caption{A dialogue example generated with the \texttt{standard} method. (Chinese Version)}
    \label{Fig-standard-dial-zh}
\end{figure*}

\begin{figure*}[t!]
    \centering
    \includegraphics[width=\textwidth]{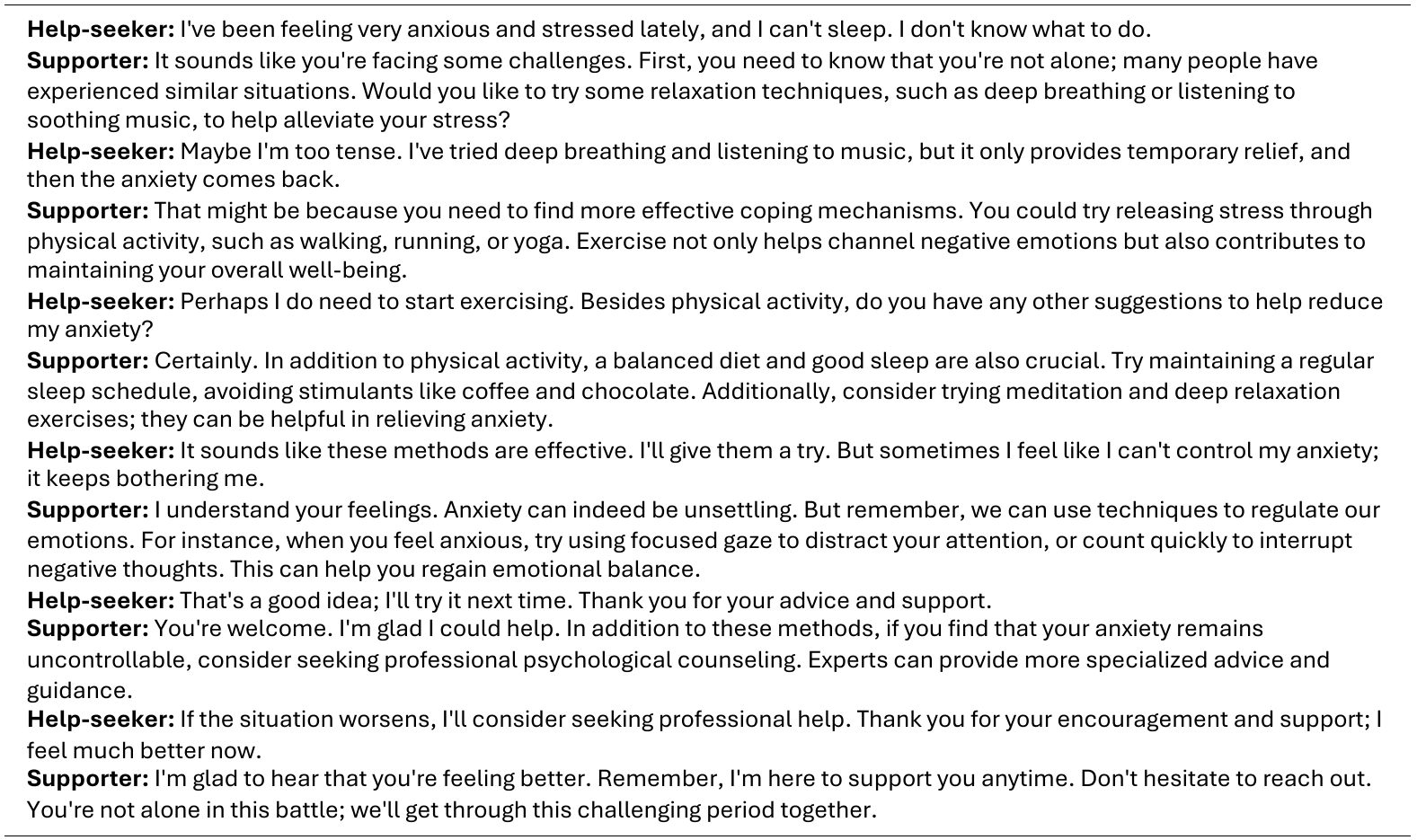}
    \caption{A dialogue example generated with the \texttt{standard} method. (English Version)}
    \label{Fig-standard-dial-en}
\end{figure*}

\begin{figure*}[t!]
    \centering
    \includegraphics[width=\textwidth]{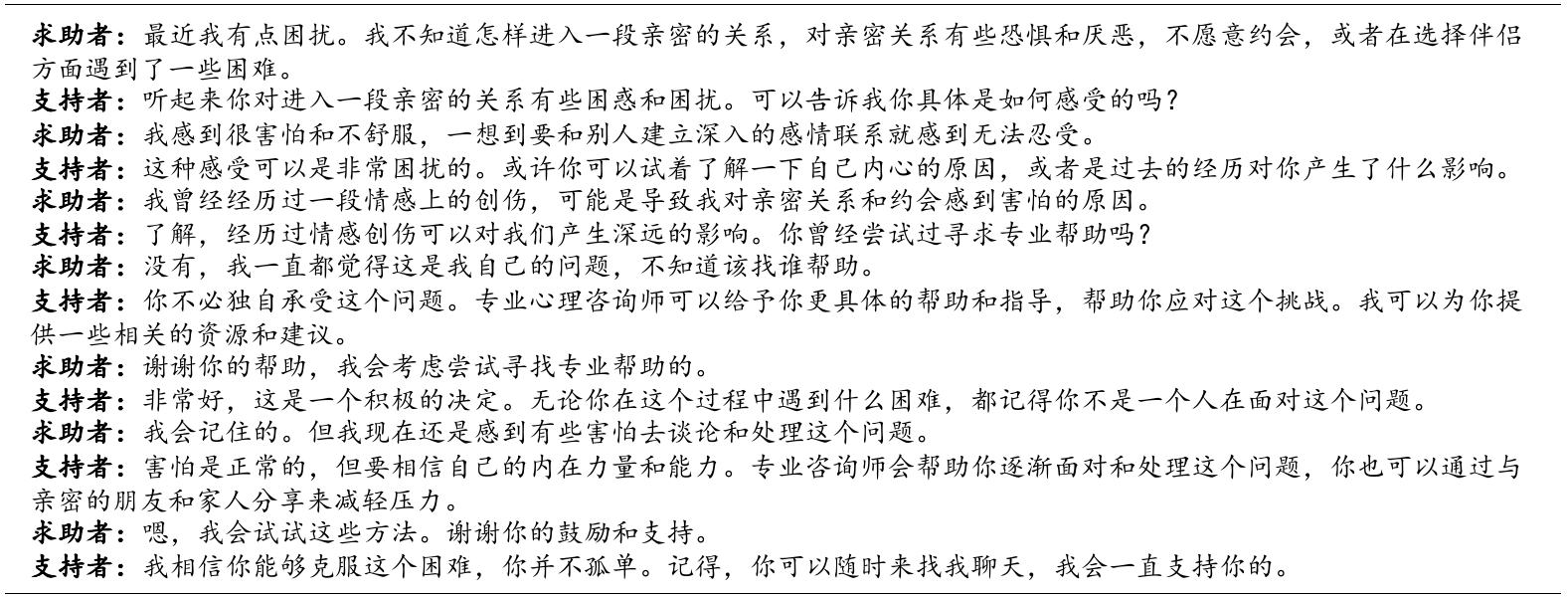}
    \caption{A dialogue example generated with the \texttt{standardT} method. (Chinese Version)}
    \label{Fig-standardT-dial-zh}
\end{figure*}

\begin{figure*}[t!]
    \centering
    \includegraphics[width=\textwidth]{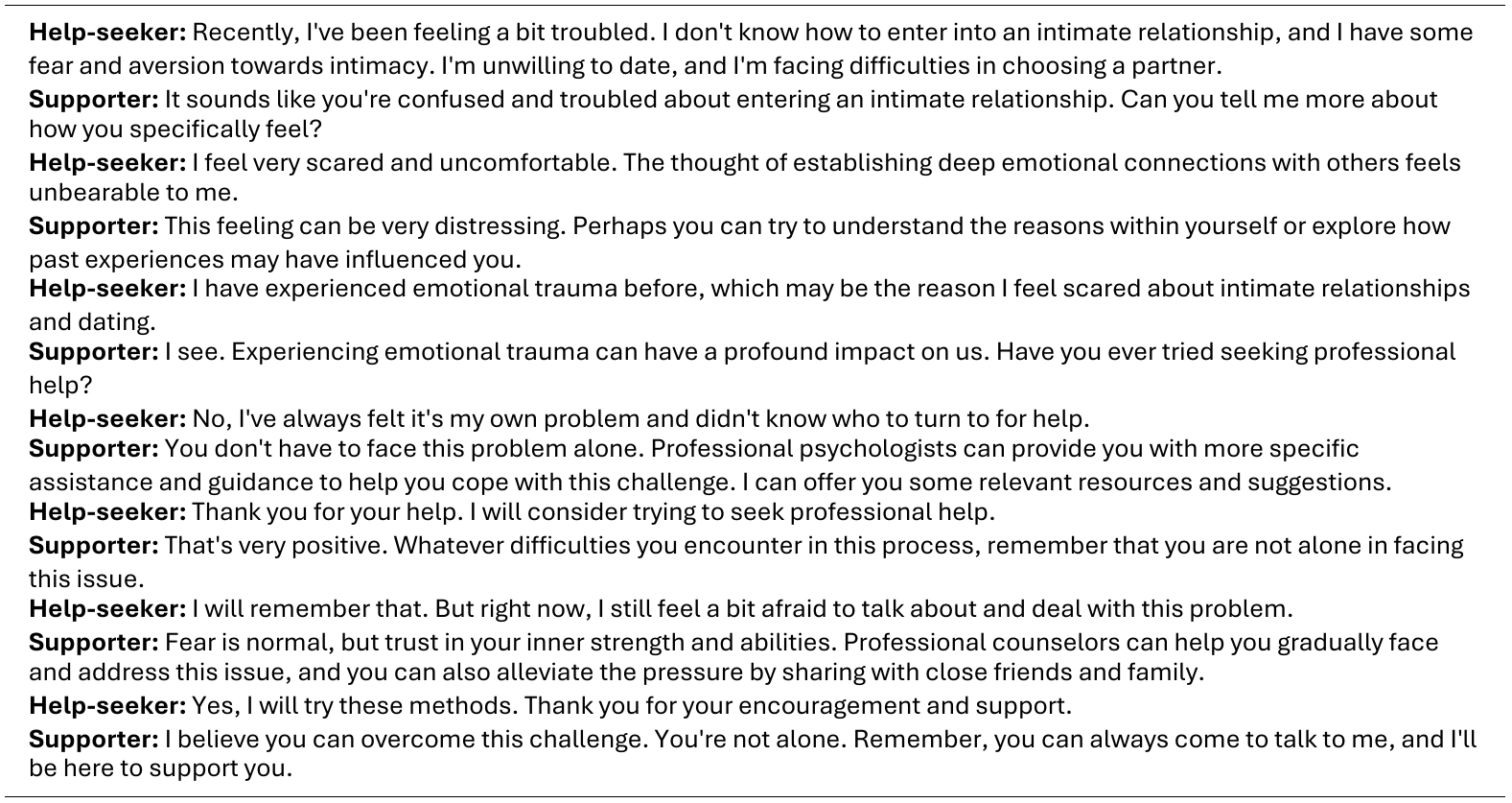}
    \caption{A dialogue example generated with the \texttt{standardT} method. (English Version)}
    \label{Fig-standardT-dial-en}
\end{figure*}

\begin{figure*}[t!]
    \centering
    \includegraphics[width=\textwidth]{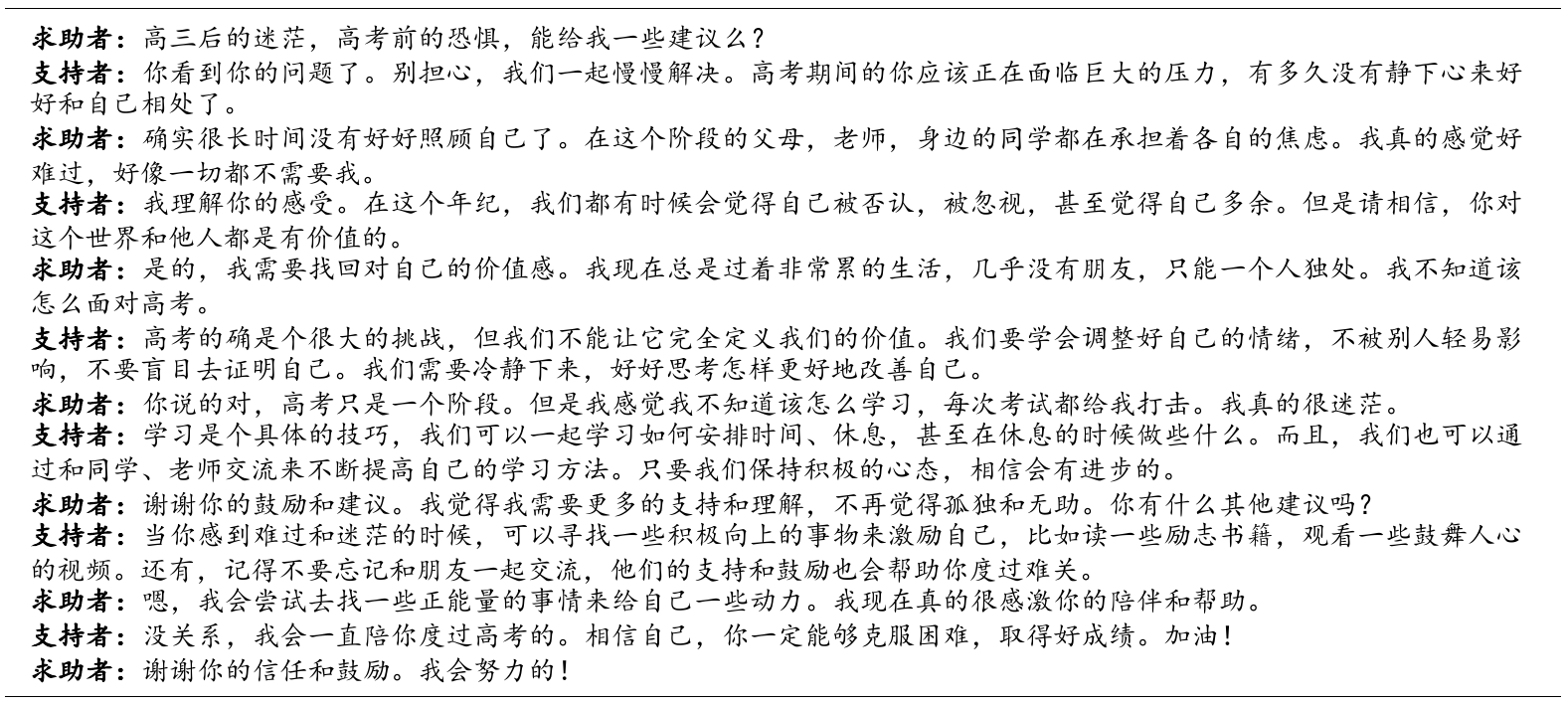}
    \caption{An example of multi-turn dialogue generated by SMILE method. (Chinese Version)}
    \label{Fig-smile-dial-zh}
\end{figure*}

\begin{figure*}[t!]
    \centering
    \includegraphics[width=\textwidth]{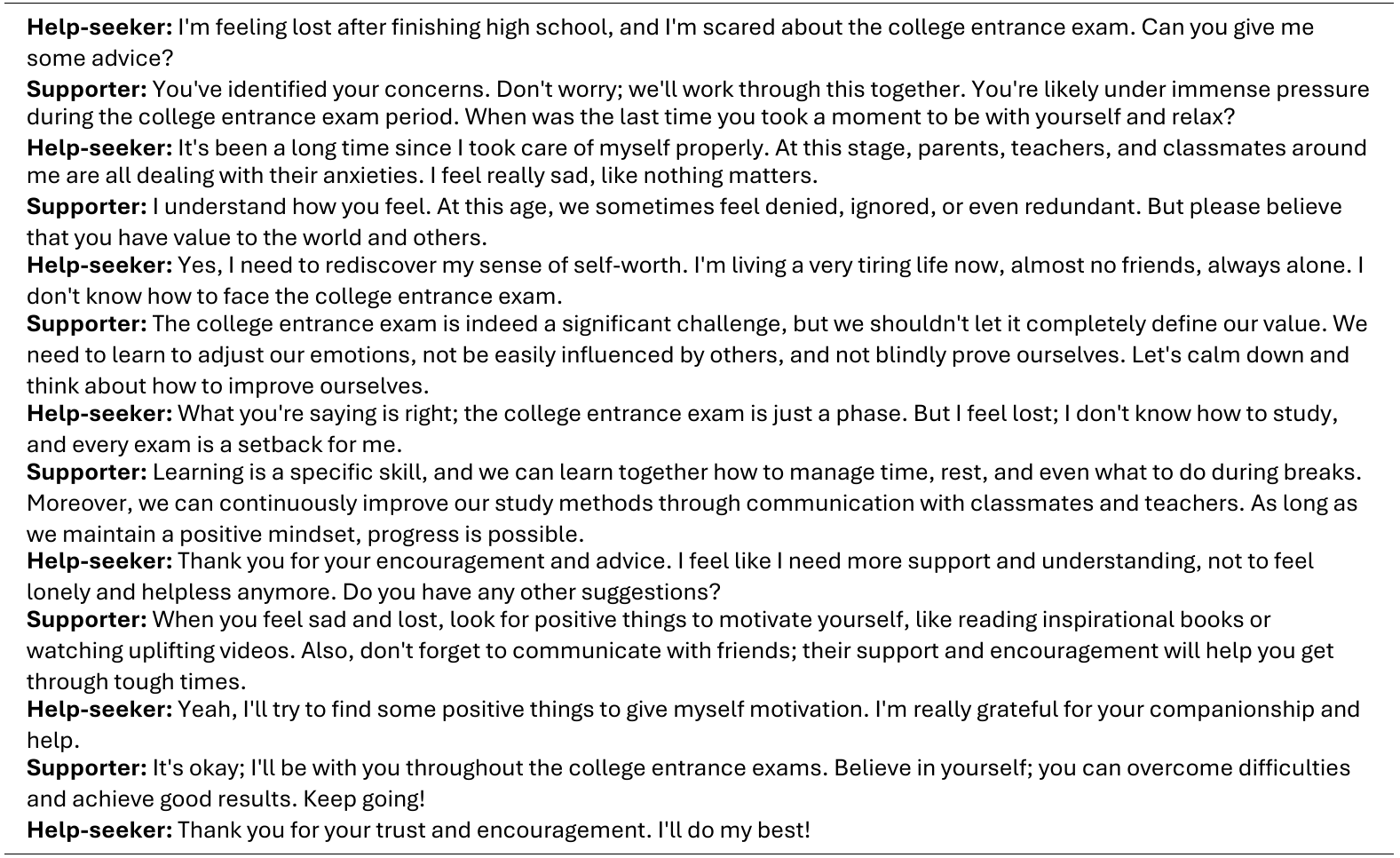}
    \caption{An example of multi-turn dialogue generated by SMILE method. (English Version)}
    \label{Fig-smile-dial-en}
\end{figure*}

Multi-turn dialogue examples generated with the \verb|standard| and \verb|standardT| methods are illustrated in Figures \ref{Fig-standard-dial-zh} and \ref{Fig-standardT-dial-zh}. Further, a dialogue generated by the SMILE method is shown in Figure \ref{Fig-smile-dial-zh}. For English version, refer to Figures \ref{Fig-standard-dial-en}, \ref{Fig-standardT-dial-en}, and \ref{Fig-smile-dial-en}.

\end{document}